\documentclass{sig-alternate-05-2015}
\usepackage{graphicx,subfigure}
\usepackage{amsmath,amssymb}
\usepackage{array}
\usepackage{color}
\usepackage{algorithmic}
\usepackage{nowidow}
\usepackage[ruled]{algorithm2e}
\definecolor{mydarkblue}{rgb}{0,0.08,0.45}
\usepackage[bookmarks, colorlinks=true, citecolor=mydarkblue,linkcolor=mydarkblue,urlcolor=mydarkblue]{hyperref}
\newcommand{\x}{\mathbf{x}}

\newcommand{\eps}{\epsilon}
\newcommand{\Rplus}{[0, +\infty)}
\newcommand{\w}{\omega}
\newcommand{\sX}{\mathcal{X}}
\newcommand{\sD}{\mathcal{D}}

\newcommand{\tdr}{\tilde{r}_{\sD}}
\newcommand{\tdrx}[1]{\tilde{r}_{\sD_{#1}}}
\newcommand{\tdw}{\tilde{\omega}_{\sD}}
\newcommand{\tdwx}[1]{\tilde{\omega}_{\sD_{#1}}}

\newcommand{\precap}{\vskip-.15in}
\newcommand{\postcap}{\vskip-.12in}

\newcommand{\presec}{\vskip-.1in}
\newcommand{\postsec}{\vskip-.01in}

\newtheorem{thm:thm}{Theorem}[section]
\newtheorem{thm:def}{Definition}[section]
\newtheorem{thm:lemma}{Lemma}[section]

\begin{document}

% Copyright
%\setcopyright{acmcopyright}
%\setcopyright{acmlicensed}
%\setcopyright{rightsretained}
%\setcopyright{usgov}
%\setcopyright{usgovmixed}
%\setcopyright{cagov}
%\setcopyright{cagovmixed}

\CopyrightYear{2016}
\setcopyright{rightsretained}
\conferenceinfo{KDD '16,}{August 13-17, 2016, San Francisco, CA, USA}
\isbn{}
\acmPrice{}
\doi{}

%% Math command

\title{XGBoost: A Scalable Tree Boosting System}

\numberofauthors{2} %  in this sample file, there are a *total*
\author{
% 1st. author
\alignauthor
Tianqi Chen\\
       \affaddr{University of Washington}\\
       \email{tqchen@cs.washington.edu}
% 2nd. author
\alignauthor
Carlos Guestrin \\
       \affaddr{University of Washington}\\
       \email{guestrin@cs.washington.edu}
}
% There's nothing stopping you putting the seventh, eighth, etc.
% author on the opening page (as the 'third row') but we ask,
% for aesthetic reasons that you place these 'additional authors'
% in the \additional authors block, viz.
\date{29 Jan. 2016}
% Just remember to make sure that the TOTAL number of pauthors
% is the number that will appear on the first page PLUS the
% number that will appear in the \additionalauthors section.

\maketitle
\begin{abstract}
Tree boosting is a highly effective and widely used machine learning method.
In this paper, we describe a scalable end-to-end tree boosting system called XGBoost, which is used widely by data scientists to achieve state-of-the-art results on many machine learning challenges. We propose a novel sparsity-aware algorithm for sparse data and weighted quantile sketch for approximate tree learning.
More importantly, we provide insights on cache access patterns, data compression and sharding to build a scalable tree boosting system.
By combining these insights, XGBoost scales beyond billions of examples using far fewer resources than existing systems.
\end{abstract}

\begin{CCSXML}
<ccs2012>
<concept_id>10010147.10010257</concept_id>
<concept_desc>Computing methodologies~Machine learning</concept_desc>
<concept_significance>500</concept_significance>
</concept>
<concept>
<concept>
<concept_id>10002951.10003227.10003351</concept_id>
<concept_desc>Information systems~Data mining</concept_desc>
<concept_significance>500</concept_significance>
</concept>
<concept>
</ccs2012>
\end{CCSXML}
% changed "Computing methodologies" to "Methodologies"
%\ccsdesc[500]{Methodologies~Machine learning}
%\ccsdesc[500]{Information systems~Data mining}


%
%  Use this command to print the description
%
\printccsdesc

% We no longer use \terms command

\keywords{Large-scale Machine Learning}

\presec
\section{Introduction}
\postsec

Machine learning and data-driven approaches are becoming very important in many areas. Smart spam classifiers protect our email by learning from massive amounts of spam data and user feedback;
advertising systems learn to match the right ads with the right context;
fraud detection systems protect banks from malicious attackers; anomaly event detection systems help experimental physicists to find events that  lead to new physics.  There are two important factors that drive these successful applications: usage of effective (statistical)~models that capture the complex data dependencies and scalable learning systems that learn the model of interest from large datasets.

Among the machine learning methods used in practice, gradient tree boosting~\cite{friedman2001greedy}\footnote{Gradient tree boosting is also known as gradient boosting machine~(GBM) or gradient boosted regression tree~(GBRT)} is one technique that shines in many applications.
Tree boosting has been shown to give state-of-the-art results on many standard classification benchmarks~\cite{Li10}.
LambdaMART~\cite{burges2010ranknet}, a variant of tree boosting for ranking, achieves state-of-the-art result for  ranking problems. Besides being used as a stand-alone predictor, it is also incorporated into real-world production pipelines for ad click through rate prediction~\cite{He:AdKDD}. Finally, it is the de-facto choice of ensemble method and is used in challenges such as the Netflix prize~\cite{bennett2007netflix}.

In this paper, we describe XGBoost, a scalable machine learning system for tree boosting. The system is available as an open source package\footnote{\url{https://github.com/dmlc/xgboost}}.
The impact of the system has been widely recognized in a number of machine learning and data mining challenges.
Take the challenges hosted by the machine learning competition site Kaggle for example.  Among the 29 challenge winning solutions~\footnote{Solutions come from of top-3 teams of each competitions.} published at Kaggle's blog during 2015, 17 solutions used XGBoost.  Among these solutions, eight solely used XGBoost to train the model, while most others combined XGBoost with neural nets in ensembles.
For comparison, the second most popular method, deep neural nets, was used in 11 solutions.
The success of the system was also witnessed in KDDCup 2015, where XGBoost was used by every winning team in the top-10.
Moreover, the winning teams reported that ensemble methods outperform a well-configured XGBoost by only a small amount ~\cite{Bekkerman:KDDCup}.

These results demonstrate that our system gives state-of-the-art results on a wide range of problems. Examples of the problems in these winning solutions include: store sales prediction; high energy physics event classification; web text classification; customer behavior prediction; motion detection; ad click through rate prediction; malware classification; product categorization; hazard risk prediction; massive online course dropout rate prediction.
While domain dependent data analysis and feature engineering play an important role in these solutions, the fact that XGBoost is the consensus choice of learner shows the impact and importance of our system and tree boosting.

The most important factor behind the success of XGBoost is its scalability in all scenarios.
The system runs more than ten times faster than existing popular solutions on a single machine and scales to billions of examples in distributed or memory-limited settings.
The scalability of XGBoost is due to several important systems and algorithmic optimizations.  These innovations include:
a novel tree learning algorithm is for handling \emph{sparse data}; a theoretically justified weighted quantile sketch procedure enables handling instance weights in
 approximate tree learning.
Parallel and distributed computing makes learning faster which enables quicker model exploration.
More importantly, XGBoost exploits out-of-core computation and enables data scientists to process hundred millions of examples on a desktop.
Finally, it is even more exciting to combine these techniques to make an end-to-end system that scales to even larger data with the least amount of cluster resources.
The major contributions of this paper is listed as follows:
\begin{itemize}
  \item We design and build a highly scalable end-to-end tree boosting system.
  \item We propose a theoretically justified weighted quantile sketch for efficient proposal calculation.
  \item We introduce a novel sparsity-aware algorithm for parallel tree learning.
  \item We propose an effective cache-aware block structure for out-of-core tree learning.
\end{itemize}

While there are some existing works on parallel tree boosting~\cite{tyree2011parallel,Ye:GBDT,PLANet},
the directions such as out-of-core computation, cache-aware and sparsity-aware learning have not been explored.
More importantly, an end-to-end system that combines all of these aspects gives a novel solution for real-world use-cases. This enables data scientists as well as researchers to build powerful variants of tree boosting algorithms~\cite{Chen:ICML2013,Chen:AISTATS2015}.
Besides these major contributions, we also make additional improvements in proposing a regularized learning objective, which we will include for completeness.

The remainder of the paper is organized as follows. We will first review tree boosting and introduce a regularized objective in Sec.~\ref{sec:model}. We then describe the split finding methods in Sec.~\ref{sec:exact} as well as the system design in Sec.~\ref{sec:system}, including experimental results when relevant to provide quantitative support for each optimization we describe. Related work is discussed in Sec.~\ref{sec:rel}.
Detailed end-to-end evaluations  are included in Sec.~\ref{sec:exp}.
Finally we conclude the paper in Sec.~\ref{sec:con}.

\presec
\section{Tree Boosting in a NutShell}\label{sec:model}
\postsec

We review gradient tree boosting algorithms in this section.
The derivation follows from the same idea in existing literatures in gradient boosting.
Specicially the second order method is originated from Friedman et al.~\cite{friedman2000additive}.
We make minor improvements in the reguralized objective,
which were found helpful in practice.

\subsection{Regularized Learning Objective}\label{subsec:obj}

\begin{figure}
\centering
\includegraphics[width=.40\textwidth]{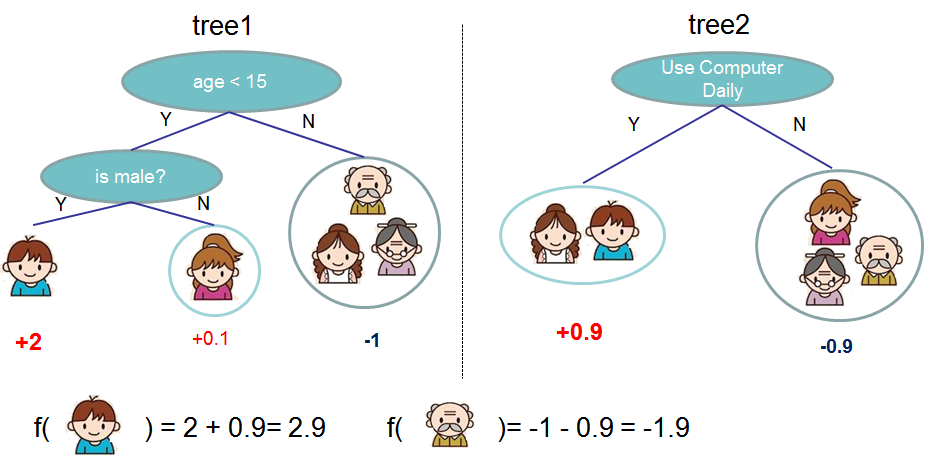}
\precap
\caption{Tree Ensemble Model. The final prediction for a given example is the sum of  predictions from each tree.}\label{fig:treemodel}
\postcap
\end{figure}

For a given data set with $n$ examples and $m$ features $\mathcal{D} = \{(\x_i, y_i)\}$ ($|\mathcal{D}| = n, \x_i\in \mathbb{R}^m, y_i \in \mathbb{R}$), a tree ensemble model~(shown in Fig.~\ref{fig:treemodel}) uses $K$ additive functions to predict the output.
\begin{equation}
\hat{y}_i = \phi(\x_i) = \sum^K_{k=1} f_k(\x_i), \ \ f_k\in \mathcal{F},
\end{equation}
where $\mathcal{F}=\{f(\x) = w_{q(\x)}\} ( q : \mathbb{R}^m \rightarrow T, w\in \mathbb{R}^T) $ is the space of regression trees (also known as CART).
Here $q$ represents the structure of each tree that maps an example to the corresponding leaf index.  $T$ is the number of leaves in the tree.
Each $f_k$ corresponds to an independent tree structure $q$ and leaf weights $w$.
Unlike decision trees, each regression tree contains a continuous score on each of the leaf, we use $w_i$ to represent score on $i$-th leaf.
For a given example, we will use the decision rules in the trees~(given by $q$) to classify it into the leaves and calculate the final prediction by summing up the score in the corresponding leaves~(given by $w$).
To learn the set of functions used in the model, we minimize the following \emph{regularized} objective.
\begin{equation}\label{eq:loss}
\begin{split}
\mathcal{L}(\phi) =& \sum_{i} l( \hat{y}_i, y_i ) + \sum_{k}\Omega( f_{k} ) \\
 \mbox{ where }& \Omega(f) = \gamma T + \frac{1}{2} \lambda \|w\|^2
 \end{split}
\end{equation}
Here $l$ is a differentiable convex loss function that measures the difference between the prediction $\hat{y}_i$ and the target $y_i$. The second term $\Omega$ penalizes the complexity of the model (i.e., the regression tree functions).
The additional regularization term helps to smooth the final learnt weights to avoid over-fitting.
Intuitively, the regularized objective will tend to select a model employing simple and predictive functions.
A similar regularization technique has been used in Regularized greedy forest~(RGF)~\cite{Tong:RGF} model.
Our objective and the corresponding learning algorithm is simpler than RGF and easier to parallelize.
When the regularization parameter is set to zero, the objective falls back to the traditional gradient tree boosting.

\subsection{Gradient Tree Boosting}
The tree ensemble model in Eq.~\eqref{eq:loss} includes functions as parameters  and cannot be optimized using  traditional optimization methods in Euclidean space.
Instead, the model is trained in an additive manner.
Formally, let $\hat{y}_i^{(t)}$ be the prediction of the $i$-th instance at the $t$-th iteration, we will need to add $f_t$ to minimize the following objective.
\begin{equation*}
\mathcal{L}^{(t)} = \sum_{i=1}^n l(y_i,\hat{y_i}^{(t-1)}+f_t(\x_i))+\Omega(f_t)
\end{equation*}
This means we greedily add the $f_t$ that most improves our model according to Eq.~\eqref{eq:loss}.
Second-order approximation can be used to quickly optimize the objective in the general setting~\cite{friedman2000additive}.
\begin{equation*}
\mathcal{L}^{(t)} \simeq \sum_{i=1}^n [l(y_i,\hat{y}^{(t-1)}) + g_i f_t(\x_i)+\frac{1}{2}h_i f_t^2(\x_i)] + \Omega(f_t)
\end{equation*}
where $g_i = \partial_{\hat{y}^{(t-1)}}l(y_i,\hat{y}^{(t-1)})$ and $h_i = \partial^2_{\hat{y}^{(t-1)}}l(y_i,\hat{y}^{(t-1)})$
are first and second order gradient statistics on the loss function.
We can remove the constant terms to obtain the following simplified objective at step $t$.
\begin{equation}\label{eq:approx}
\tilde{\mathcal{L}}^{(t)} = \sum_{i=1}^n [g_i f_t(\x_i)+\frac{1}{2}h_i f_t^2(\x_i)] + \Omega(f_t)
\end{equation}

Define $I_j=\{i|q(\x_i)=j\}$ as the instance set of leaf $j$.
We can rewrite Eq~\eqref{eq:approx} by expanding $\Omega$ as follows
\begin{equation}
\begin{split}
\tilde{\mathcal{L}}^{(t)}
         &=\sum^n_{i=1} [g_i f_t(\x_i)+\frac{1}{2}h_if_t^2(\x_i)]+\gamma T + \frac{1}{2}\lambda\sum^T_{j=1}w_j^2\\
         &=\sum^T_{j=1}[(\sum_{i\in I_j} g_i)w_j+\frac{1}{2}(\sum_{i\in I_j} h_i+\lambda)w_j^2]+\gamma T
\end{split}
\end{equation}

For a fixed structure $q(\x)$, we can compute the optimal weight $w_j^*$ of leaf $j$ by
\begin{equation}\label{eq:leafscore}
w^*_j =-\frac{\sum_{i\in I_j} g_i}{\sum_{i\in I_j} h_i+\lambda},
\end{equation}
and calculate the corresponding optimal  value by
\begin{equation}\label{eq:score}
\tilde{\mathcal{L}}^{(t)}(q) = - \frac{1}{2} \sum^T_{j=1}\frac{(\sum_{i\in I_j} g_i)^2}{\sum_{i\in I_j} h_i + \lambda}+\gamma T.
\end{equation}

\begin{figure}
\centering
\includegraphics[width=.40\textwidth]{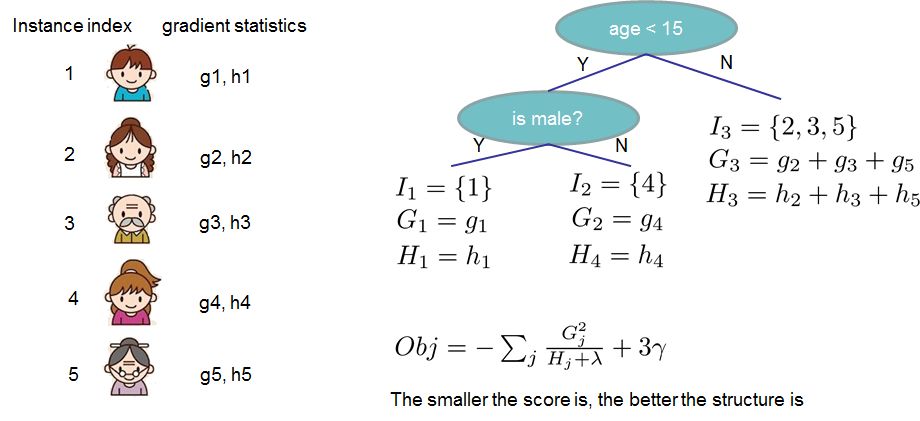}
\precap
\caption{Structure Score Calculation. We only need to sum up the gradient and second order gradient statistics on each leaf, then apply the scoring formula to get the quality score.}\label{fig:structscore}
\postcap
\end{figure}

Eq~\eqref{eq:score} can be used as a scoring function to measure the quality of a tree structure $q$. This score is like the impurity score for evaluating decision trees, except that it is derived for a wider range of objective functions.
Fig.~\ref{fig:structscore} illustrates how this score can be calculated.

Normally it is impossible to enumerate all the possible tree structures $q$.
A greedy algorithm that starts from a single leaf and iteratively adds branches to the tree is used instead.
Assume that $I_L$ and $I_R$ are the instance sets of left and right nodes after the split.
Lettting $I= I_L\cup I_R$, then the loss reduction after the split is given by
\begin{equation}\label{eq:gain}
      \mathcal{L}_{split} =\frac{1}{2} \left[\frac{(\sum_{i\in I_L} g_i)^2}{\sum_{i\in I_L} h_i + \lambda}+\frac{(\sum_{i\in I_R} g_i)^2}{\sum_{i\in I_R} h_i + \lambda} - \frac{(\sum_{i\in I} g_i)^2}{\sum_{i\in I} h_i + \lambda}\right] - \gamma
\end{equation}
This formula is usually used in practice for evaluating the split candidates.

\subsection{Shrinkage and Column Subsampling}
Besides the regularized objective mentioned in Sec.~\ref{subsec:obj}, two additional techniques are used to further prevent over-fitting.
The first technique is shrinkage introduced by Friedman~\cite{friedman2002stochastic}.
Shrinkage scales newly added weights by a factor $\eta$ after each step of tree boosting.
Similar to a learning rate in tochastic optimization, shrinkage reduces the influence of each individual tree and leaves space for future trees to improve the model.
The second technique is column (feature) subsampling. This technique is used in RandomForest~\cite{Breiman:RF,Friedman03importancesampled},
It is implemented in a commercial software TreeNet~\footnote{https://www.salford-systems.com/products/treenet} for gradient boosting,
but is not implemented in existing opensource packages.
According to user feedback, using column sub-sampling prevents over-fitting even more so than the traditional row sub-sampling (which is also supported). The usage of column sub-samples also speeds up computations of the parallel algorithm described later.
\presec
\section{Split Finding Algorithms}\label{sec:exact}
\postsec

\subsection{Basic Exact Greedy Algorithm}
\begin{algorithm}[t]
    \caption{Exact Greedy Algorithm for Split Finding}\label{alg:exact-greedy}
    \KwIn{$I$, instance set of current node}
    \KwIn{$d$, feature dimension}

    $gain\leftarrow 0$\\
    $G \leftarrow \sum_{i\in I} g_{i}$, $H \leftarrow \sum_{i \in I} h_{i}$\\
    \For{$k=1$ {\bfseries to} $m$ }{
       $G_L\leftarrow 0,\ H_L\leftarrow 0$\\
       \For{$j $ in sorted($I$, by $\x_{jk}$)}{
            $G_L\leftarrow G_L + g_j,\ H_L\leftarrow H_L + h_j$\\
            $G_R\leftarrow G - G_L,\ H_R\leftarrow H - H_L$\\
            $score \leftarrow \max(score, \frac{G_{L}^2}{H_{L}+\lambda} + \frac{G_{R}^2}{H_{R}+\lambda} - \frac{G^2}{H+\lambda})$\\
       }
    }
    \KwOut{Split with max score}
\end{algorithm}

One of the key problems in tree learning is to find the best split as indicated by Eq~\eqref{eq:gain}.
In order to do so, a split finding algorithm enumerates over all the possible splits on all the features. We call this the \emph{exact greedy algorithm}.
Most existing single machine tree boosting implementations, such as scikit-learn~\cite{scikit-learn}, R's gbm~\cite{RGBM} as well as the single machine version of XGBoost support the exact greedy algorithm.
The exact greedy algorithm is shown in Alg.~\ref{alg:exact-greedy}.
It is computationally demanding to enumerate all the possible splits for continuous features.
In order to do so efficiently, the algorithm must first  sort the data according to feature values and visit the data in sorted order to accumulate the gradient statistics
for the structure score in Eq~\eqref{eq:gain}.

\subsection{Approximate Algorithm}

\begin{algorithm}[t]
    \caption{Approximate Algorithm for Split Finding}\label{alg:approx-greedy}
     \For{$k=1$ {\bfseries to} $m$ }{
        \mbox{Propose $S_k = \{s_{k1}, s_{k2}, \cdots s_{kl}\}$ by percentiles on feature $k$.}\\
        \mbox{Proposal can be done per tree~(global), or per split(local).}
     }
     \For{$k=1$ {\bfseries to} $m$ }{
        $G_{kv} \leftarrow = \sum_{j\in \{j| s_{k, v}\geq \x_{jk} > s_{k, v-1}\}} g_j$
        $H_{kv} \leftarrow = \sum_{j\in \{j| s_{k, v}\geq \x_{jk} > s_{k, v-1}\}} h_j$
     }
     Follow same step as in previous section to find max score only among proposed splits.\\
\end{algorithm}

The exact greedy algorithm is very powerful since it enumerates over all possible splitting points greedily.
However, it is impossible to efficiently  do so when the data does not fit entirely into memory.
Same problem also arises in the distributed setting.
To support effective gradient tree boosting in these two settings, an approximate algorithm is needed.

We summarize an approximate framework, which resembles the ideas proposed in past literatures~\cite{McRank,Bekkerman:Scale,tyree2011parallel},
in Alg.~\ref{alg:approx-greedy}.
To summarize, the algorithm  first proposes candidate splitting points according to percentiles of feature distribution~(a specific criteria will be given in Sec.~\ref{sec:quantile}).
The algorithm then maps the continuous features into buckets split by these candidate points, aggregates the statistics and finds the best solution among proposals based on the aggregated statistics.

\begin{figure}
    \centering
  \includegraphics[width=.36\textwidth]{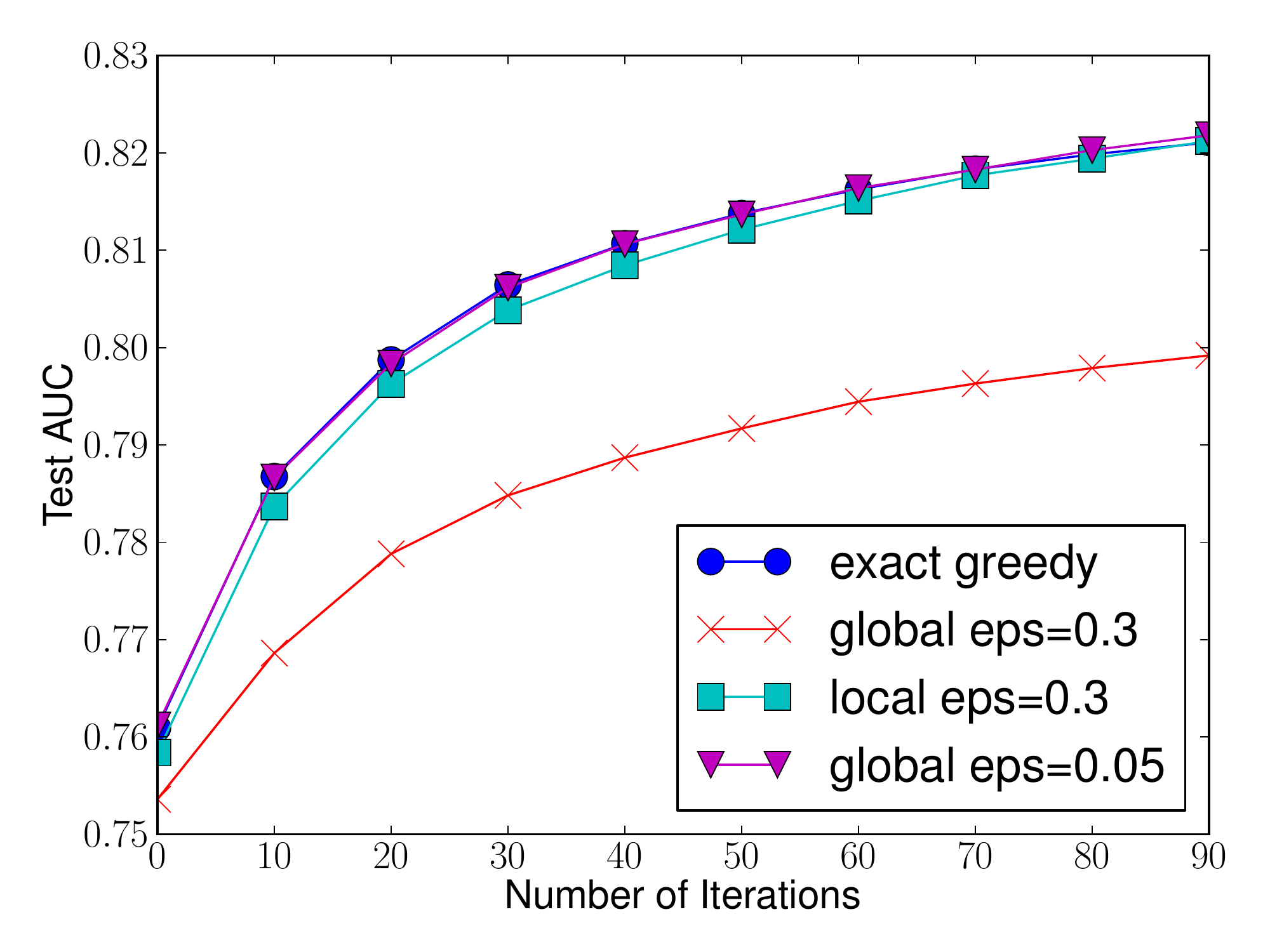}
\precap
  \caption{Comparison of test AUC convergence on Higgs 10M dataset.
The eps parameter corresponds to the accuracy of the approximate sketch. This roughly translates to 1 / eps buckets in the proposal.
We  find that local proposals require fewer buckets, because it refine split candidates.
  }\label{fig:higgs-approx-cmp-10m}

\end{figure}

There are two variants of the algorithm, depending on when the proposal is given.
The global variant  proposes all the candidate splits during the initial phase of tree construction,
and uses the same proposals for split finding at all levels. The local variant  re-proposes after each split.
The global method requires less proposal steps than the local method. However, usually more candidate points are needed for the global proposal
 because candidates are not refined after each split.
The local proposal refines the candidates after splits, and can potentially be more appropriate for deeper trees.
A comparison of different algorithms on a Higgs boson dataset is given by Fig.~\ref{fig:higgs-approx-cmp-10m}.
We find that the local proposal indeed requires fewer candidates. The global proposal can be as accurate as the local one given enough candidates.

Most existing approximate algorithms for distributed tree learning also follow this framework.
Notably, it is also possible to directly construct approximate histograms of gradient statistics~\cite{tyree2011parallel}.
It is also possible to use other variants of binning strategies instead of quantile~\cite{McRank}.
Quantile strategy benefit from being distributable and recomputable, which we will detail in next subsection.
From Fig.~\ref{fig:higgs-approx-cmp-10m}, we also find that the quantile strategy can get the same accuracy as exact greedy given reasonable approximation level.

Our system efficiently supports exact greedy for the single machine setting, as well as approximate algorithm with both local and global proposal methods for all settings.
Users can freely choose between the methods according to their needs.

\subsection{Weighted Quantile Sketch}\label{sec:quantile}
One important step in the approximate algorithm is to propose candidate split points. Usually percentiles of a feature are used to make candidates distribute evenly on the data.
Formally, let multi-set
$\sD_k=\{(x_{1k}, h_1), (x_{2k}, h_2) \cdots (x_{nk}, h_n)\}$
represent the $k$-th feature values and second order gradient statistics of each training instances.
We can define a rank functions $r_{k} : \mathbb{R} \rightarrow \Rplus$ as
\begin{equation}
    r_{k}(z) =\frac{1}{\sum_{(x, h)\in \sD_k} h} \sum_{(x, h)\in \sD_k, x < z} h,
\end{equation}
which represents the proportion of instances whose feature value $k$ is smaller than $z$.
The goal is to find candidate split points $\{s_{k1}, s_{k2}, \cdots s_{kl}\}$, such that
\begin{equation}
    |r_{k}(s_{k,j})  - r_{k}(s_{k,j+1})| < \eps, \ \ s_{k1} = \min_i \x_{ik},  s_{kl} = \max_i \x_{ik}.
\end{equation} \label{eq:quantile}
Here $\eps$ is an approximation factor.
Intuitively, this means that there is roughly $1 / \eps$ candidate points.
Here each data point is weighted by $h_i$.
To see why $h_i$ represents the weight, we can rewrite  Eq~\eqref{eq:approx} as
$$
\sum_{i=1}^n \frac{1}{2} h_i (f_t(\x_i) - g_i / h_i )^2 + \Omega(f_t) + constant,
$$
which is exactly weighted squared loss with labels $g_i/h_i$ and weights $h_i$.
For large datasets, it is non-trivial to find candidate splits that satisfy the criteria.
When every instance has equal  weights, an existing algorithm called quantile sketch~\cite{Greenwald:SIGMOID01,Zhang:SSDBM}
solves the problem. However, there is no existing quantile sketch for the weighted datasets.
Therefore, most existing approximate algorithms either resorted to sorting on a random subset of data which have a chance of failure or heuristics that do not have theoretical guarantee.

To solve this problem, we introduced a novel distributed weighted quantile sketch algorithm that can handle weighted data with a
\emph{provable theoretical guarantee}. The general idea is to propose a data structure that supports \emph{merge} and \emph{prune}
operations, with each operation  proven to maintain a certain accuracy level.
A detailed description of the algorithm as well as proofs are given in the appendix.

\subsection{Sparsity-aware Split Finding} \label{subsec:sparse}

\begin{algorithm}[!t]
    \caption{Sparsity-aware Split Finding}\label{alg:sparse-split}
    \KwIn{$I$, instance set of current node}
    \KwIn{$I_k =\{i\in I|x_{ik} \neq \mbox{missing}\}$}
    \KwIn{$d$, feature dimension}
    \emph{Also applies to the approximate setting, only collect statistics of non-missing entries into buckets}\\
    $gain\leftarrow 0$\\
    $G \leftarrow \sum_{i\in I},  g_{i}$,$H \leftarrow \sum_{i \in I} h_{i}$\\
    \For{$k=1$ {\bfseries to} $m$}{
       \emph{// enumerate missing value goto right}\\
       $G_L\leftarrow 0,\ H_L\leftarrow 0$\\
       \For{$j $ in sorted($I_k$, ascent order by $\x_{jk}$)}{
            $G_L\leftarrow G_L + g_j,\ H_L\leftarrow H_L + h_j$\\
            $G_R\leftarrow G - G_L,\ H_R\leftarrow H - H_L$\\
            $score \leftarrow \max(score, \frac{G_{L}^2}{H_{L}+\lambda} + \frac{G_{R}^2}{H_{R}+\lambda} - \frac{G^2}{H+\lambda})$\\
       }
       \emph{// enumerate missing value goto left}\\
       $G_R\leftarrow 0,\ H_R\leftarrow 0$\\
       \For{$j $ in sorted($I_k$, descent order by $\x_{jk}$)}{
            $G_R\leftarrow G_R + g_j,\ H_R\leftarrow H_R + h_j$\\
            $G_L\leftarrow G - G_R,\ H_L\leftarrow H - H_R$\\
            $score \leftarrow \max(score, \frac{G_{L}^2}{H_{L}+\lambda} + \frac{G_{R}^2}{H_{R}+\lambda} - \frac{G^2}{H+\lambda})$\\
       }
    }
    \KwOut{Split and default directions with max gain}
\end{algorithm}

\begin{figure}[!t]
\centering
\includegraphics[width=.35\textwidth]{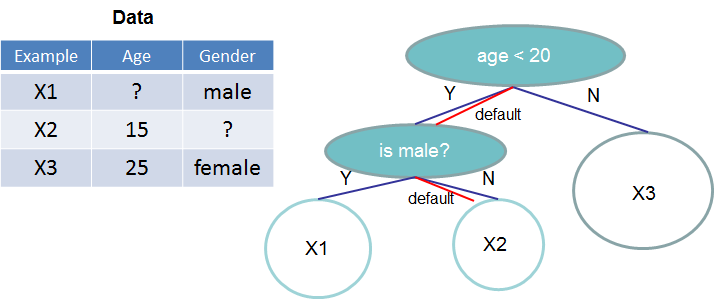}
\caption{Tree structure with default directions.
    An example will be classified into the default direction when the feature needed for the split is missing.
}\label{fig:default}
\end{figure}
\begin{figure}[t]
  \centering
  \includegraphics[width=.33\textwidth]{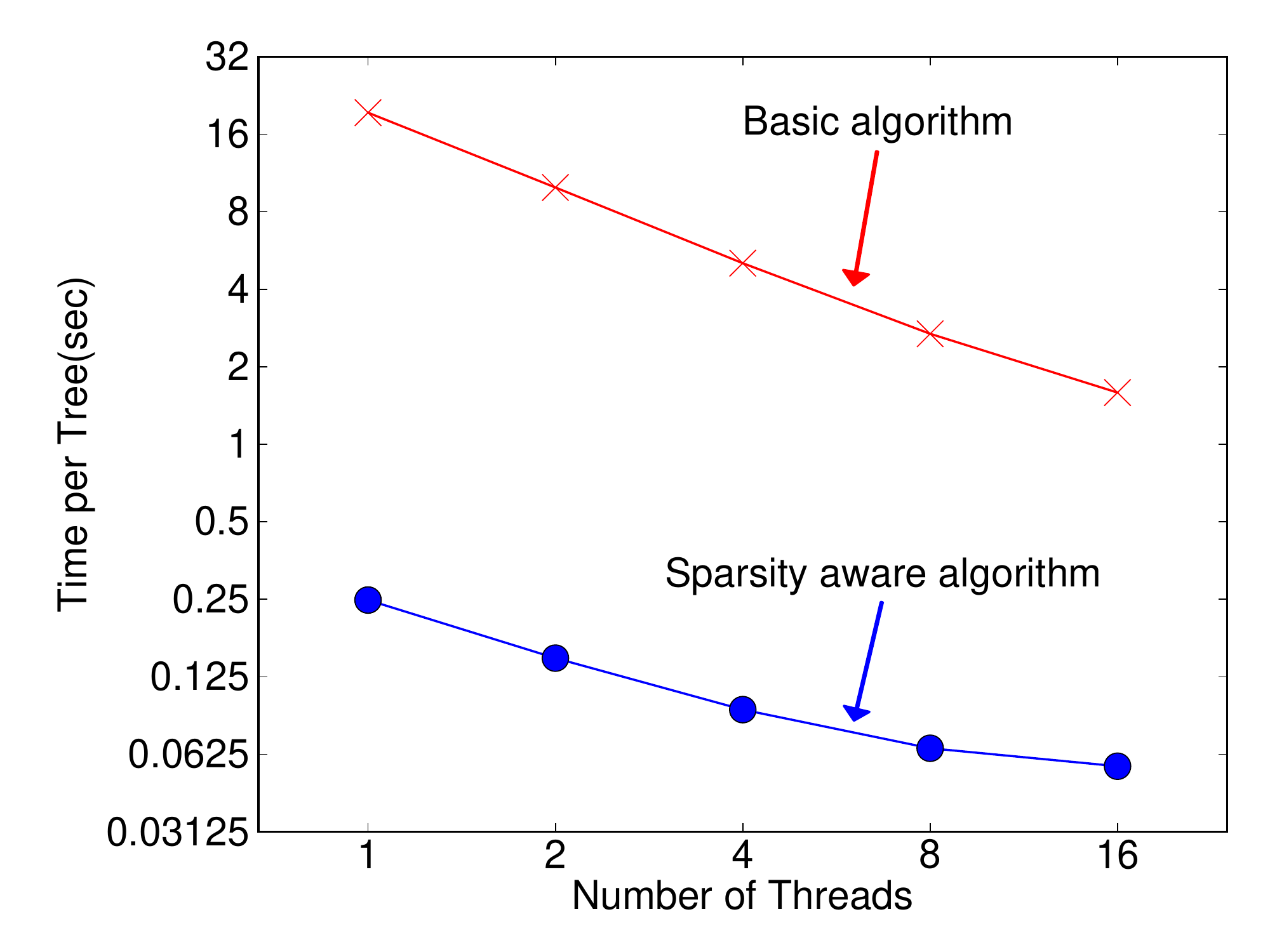}
\precap
  \caption{Impact of the sparsity aware algorithm on Allstate-10K.
    The dataset is sparse mainly due to one-hot encoding. The sparsity aware algorithm is more than 50 times faster than the naive version that does not take sparsity into consideration.
  }\label{fig:sparse-allstate}
\postcap
\end{figure}

\begin{figure*}
\centering
\includegraphics[width=.8\textwidth]{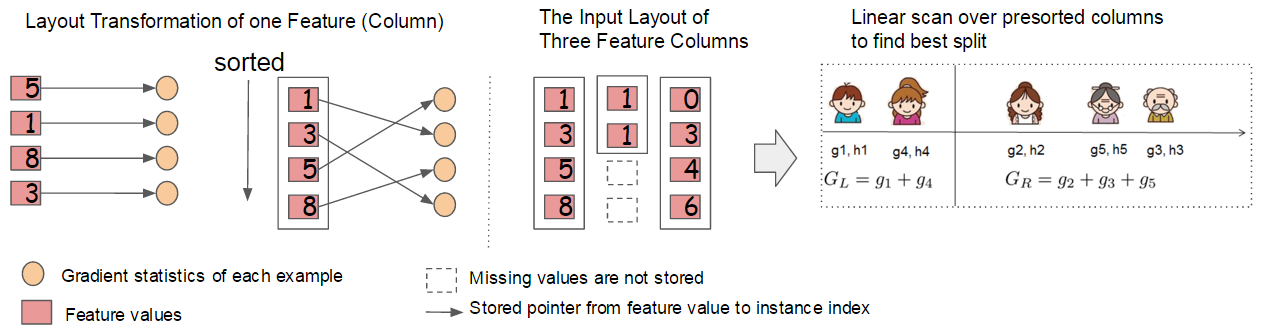}
\precap
\caption{Block structure for  parallel learning. Each column in a block is sorted by the corresponding feature value.
A linear scan over one column in the block is sufficient to enumerate all the split points.}\label{fig:layout}
\postcap
\end{figure*}

In many real-world problems, it is quite common for the input $\x$ to be sparse.
There are multiple possible causes for sparsity: 1) presence of missing values in the data;
2) frequent zero entries in the statistics; and, 3) artifacts of feature engineering such as one-hot encoding.
It is important to make the algorithm aware of the sparsity pattern in the data.
In order to do so, we propose to add a default direction in each tree node, which is shown in Fig.~\ref{fig:default}.
When a value is missing in the sparse matrix $\x$, the instance is classified into the default direction.
There are two choices of default direction in each branch. The optimal default directions are learnt from the data.
The algorithm is shown in Alg.~\ref{alg:sparse-split}. The key improvement is to only visit the non-missing entries $I_k$. The presented algorithm treats the non-presence as a missing value and learns the best direction to handle missing values.
The same algorithm can also be applied when the non-presence corresponds to a user specified value by limiting the enumeration only to consistent solutions.

To the best of our knowledge, most existing tree learning algorithms are either only optimized for dense data, or need specific procedures to handle limited cases such as categorical encoding.
XGBoost handles all sparsity patterns in a unified way.
More importantly, our method exploits the sparsity to make computation complexity linear to number of non-missing entries in the input.
Fig.~\ref{fig:sparse-allstate} shows the comparison of sparsity aware and a naive implementation on an Allstate-10K dataset~(description of dataset  given in Sec.~\ref{sec:exp}). We find that the sparsity aware algorithm runs 50 times faster than the naive version.
This confirms the importance of the sparsity aware algorithm.

\begin{figure*}[!ct]
  \centering
  \subfigure[Allstate 10M] {
    \includegraphics[width=.23\textwidth]{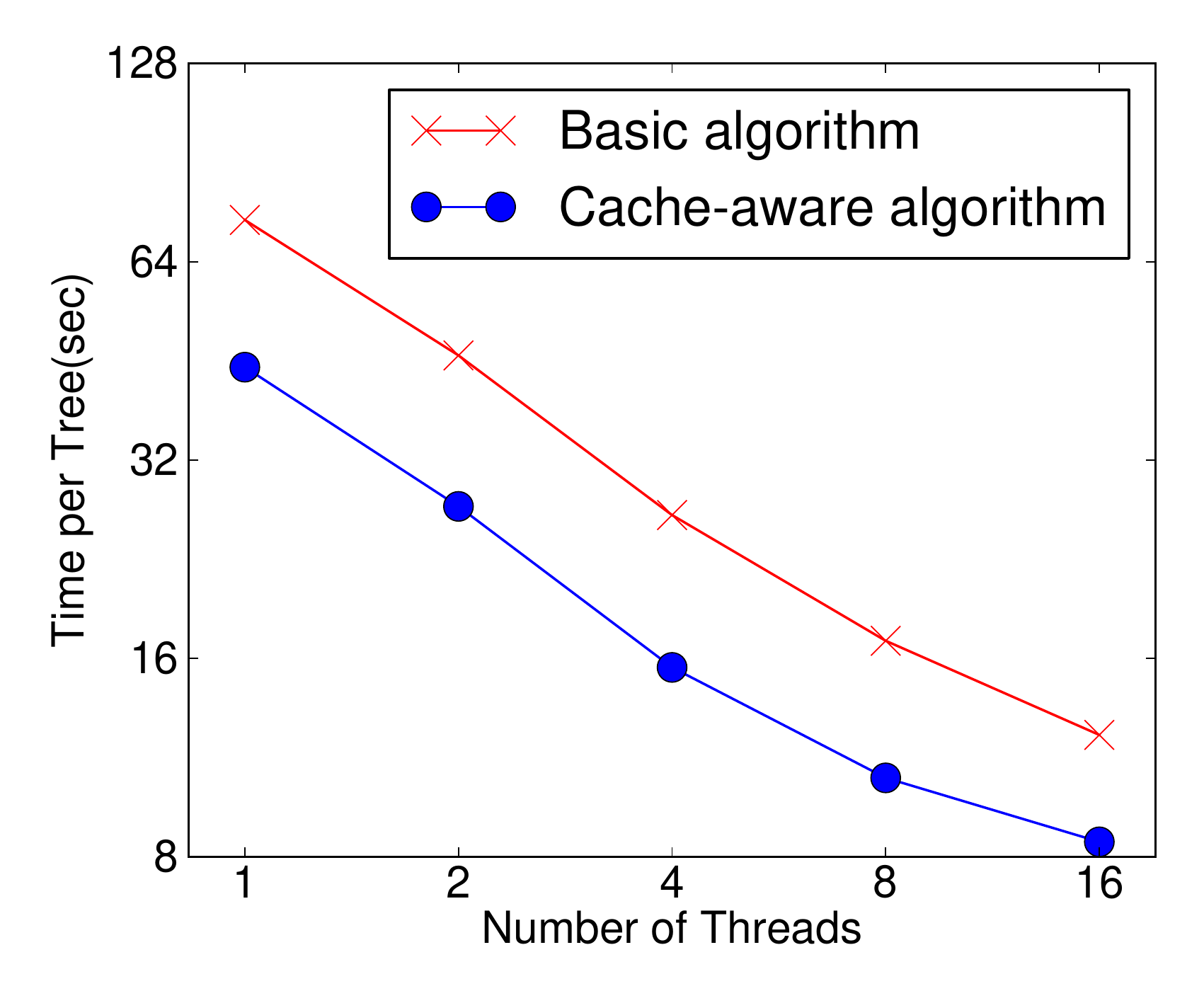}
  }
  \subfigure[Higgs 10M] {
    \includegraphics[width=.23\textwidth]{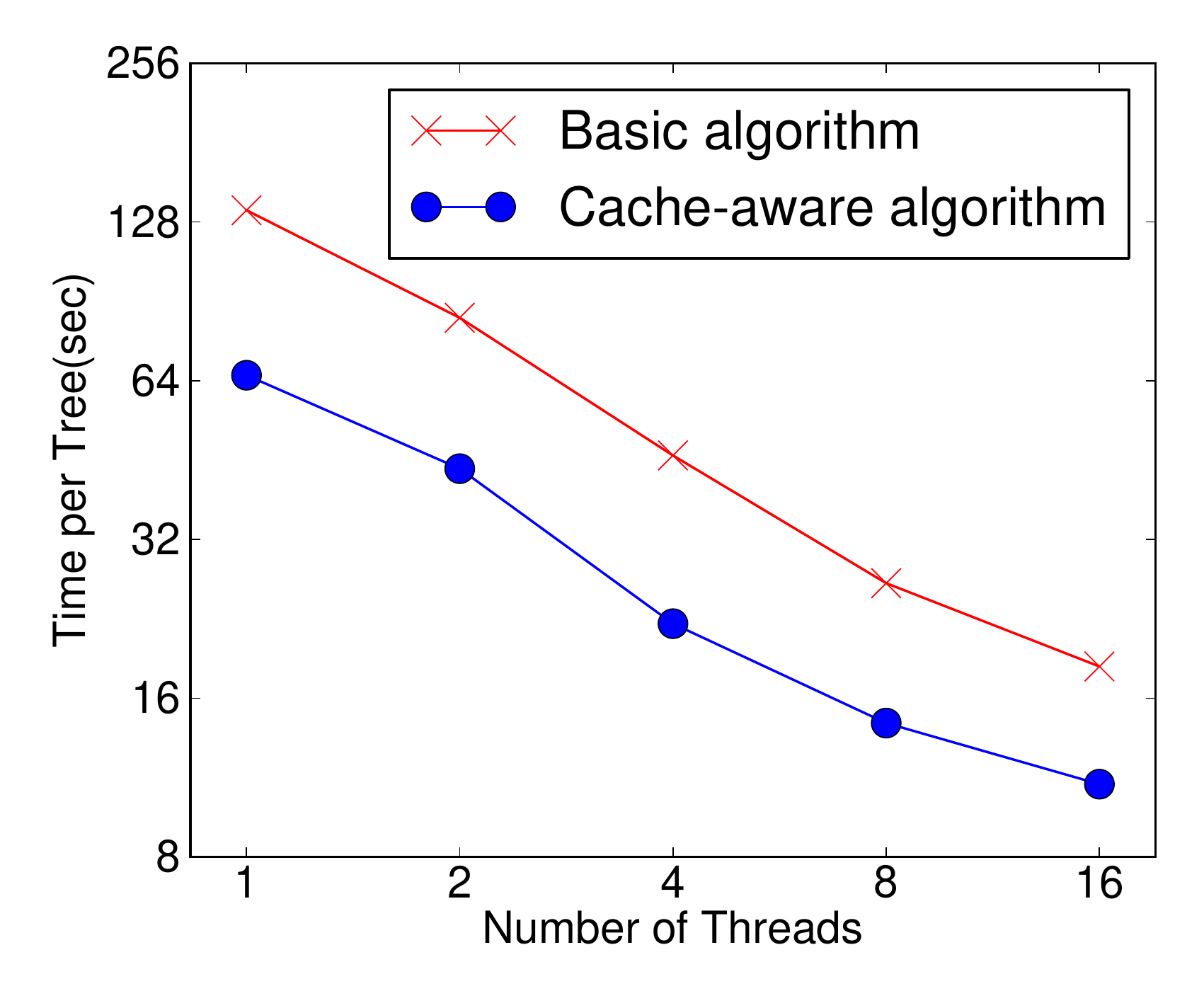}
  }
  \subfigure[Allstate 1M] {
    \includegraphics[width=.23\textwidth]{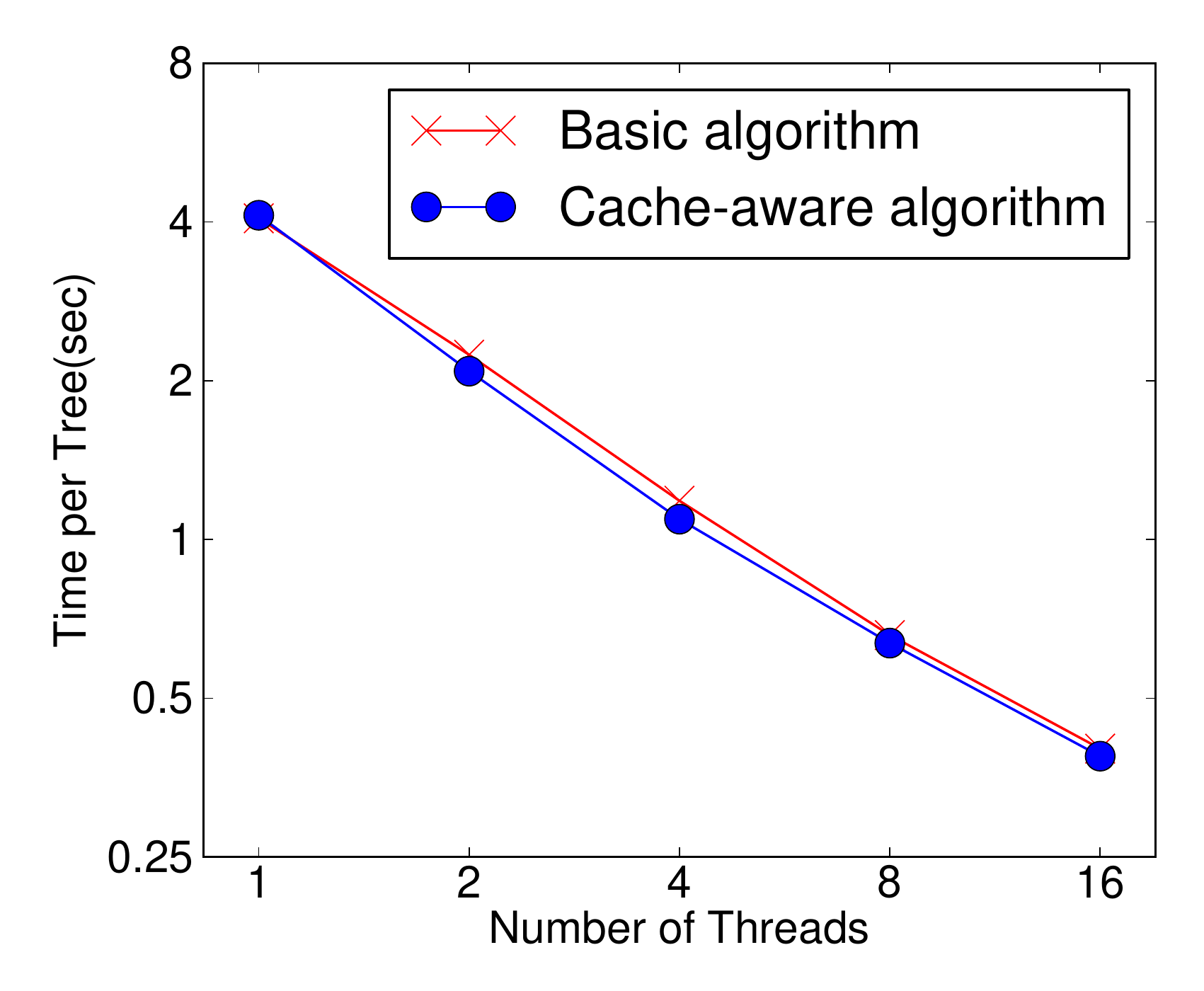}
  }
  \subfigure[Higgs 1M] {
    \includegraphics[width=.23\textwidth]{higgs-cache-bynthread-1m}
  }
    \precap
  \caption{Impact of cache-aware prefetching in exact greedy algorithm.
    We find that the cache-miss effect impacts the performance on the large datasets~(10 million instances).
    Using cache aware prefetching improves the performance by factor of two when the dataset is large.
  }\label{fig:exact-cache}
    \postcap
\end{figure*}

\presec
\section{System Design} \label{sec:system}
\postsec

\subsection{Column Block for Parallel Learning}
The most time consuming part of  tree learning  is to get the data into sorted order.
In order to reduce the cost of sorting, we propose to store the data in in-memory units,  which we called \emph{block}.
Data in each block is stored in the compressed column~(CSC) format, with each column sorted by the corresponding feature value.
This input data layout only needs to be computed once before training, and can be reused in later iterations.

In the exact greedy algorithm, we store the entire dataset in a single block and run the split search algorithm by linearly scanning over the pre-sorted entries. We do the split finding of all leaves collectively, so one scan over the block will collect the statistics of the split candidates in all leaf branches. Fig.~\ref{fig:layout} shows how we transform a dataset into the format and find the optimal split using the block structure.

The block structure also helps when using the approximate algorithms.
Multiple blocks can be used in this case, with each block corresponding to subset of rows in the dataset.
Different blocks can be distributed across machines, or stored on disk in the out-of-core setting.
Using the sorted structure, the quantile finding step becomes a \emph{linear scan} over the sorted columns.
This is especially valuable for local proposal algorithms, where candidates are generated frequently at each branch.
The binary search in histogram aggregation also becomes  a linear time merge style algorithm.

Collecting statistics for each column can be \emph{parallelized}, giving us a parallel algorithm for split finding.
Importantly, the column block structure also supports column subsampling, as it is easy to select a subset of columns in a block.

\noindent \textbf{Time Complexity Analysis}
Let $d$ be the maximum depth of the tree and $K$ be total number of trees.
For the exact greedy algorithm, the time complexity of original spase aware algorithm is $O(K d \|\x\|_0 \log n)$.
Here we use $\|\x\|_0$ to denote number of non-missing entries in the training data.
On the other hand, tree boosting on the block structure only cost $O(K d \|\x\|_0 + \|\x\|_0 \log n)$.
Here $O( \|\x\|_0 \log n)$ is the one time preprocessing cost that can be amortized.
This analysis shows that the block structure helps to save an additional $\log n$ factor, which is significant when $n$ is large.
For the approximate algorithm, the time complexity of original algorithm with binary search is $O(K d \|\x\|_0 \log q)$.
Here $q$ is the number of proposal candidates in the dataset. While $q$ is usually between 32 and 100, the log factor still introduces overhead.
Using the block structure, we can reduce the time to $O(K d \|\x\|_0 + \|\x\|_0 \log B)$, where $B$ is the maximum number of rows in each block.
Again we can save the additional $\log q$ factor in computation.

\subsection{Cache-aware Access}

\begin{figure}[!t]
  \centering
  \includegraphics[width=.34\textwidth]{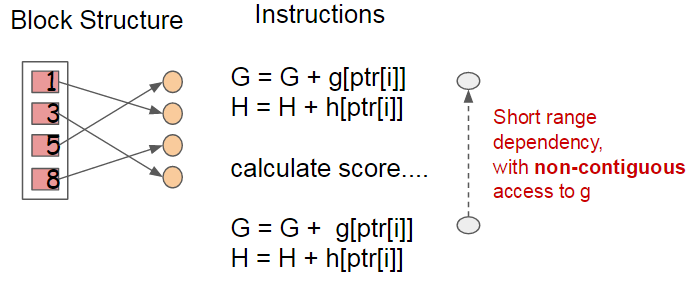}
\precap
  \caption{
    Short range data dependency pattern that can cause stall due to cache miss.
  } \label{fig:cache-miss}
\postcap
\end{figure}
\begin{figure}[!t]
  \centering
  \subfigure[Allstate 10M] {
    \includegraphics[width=.28\textwidth]{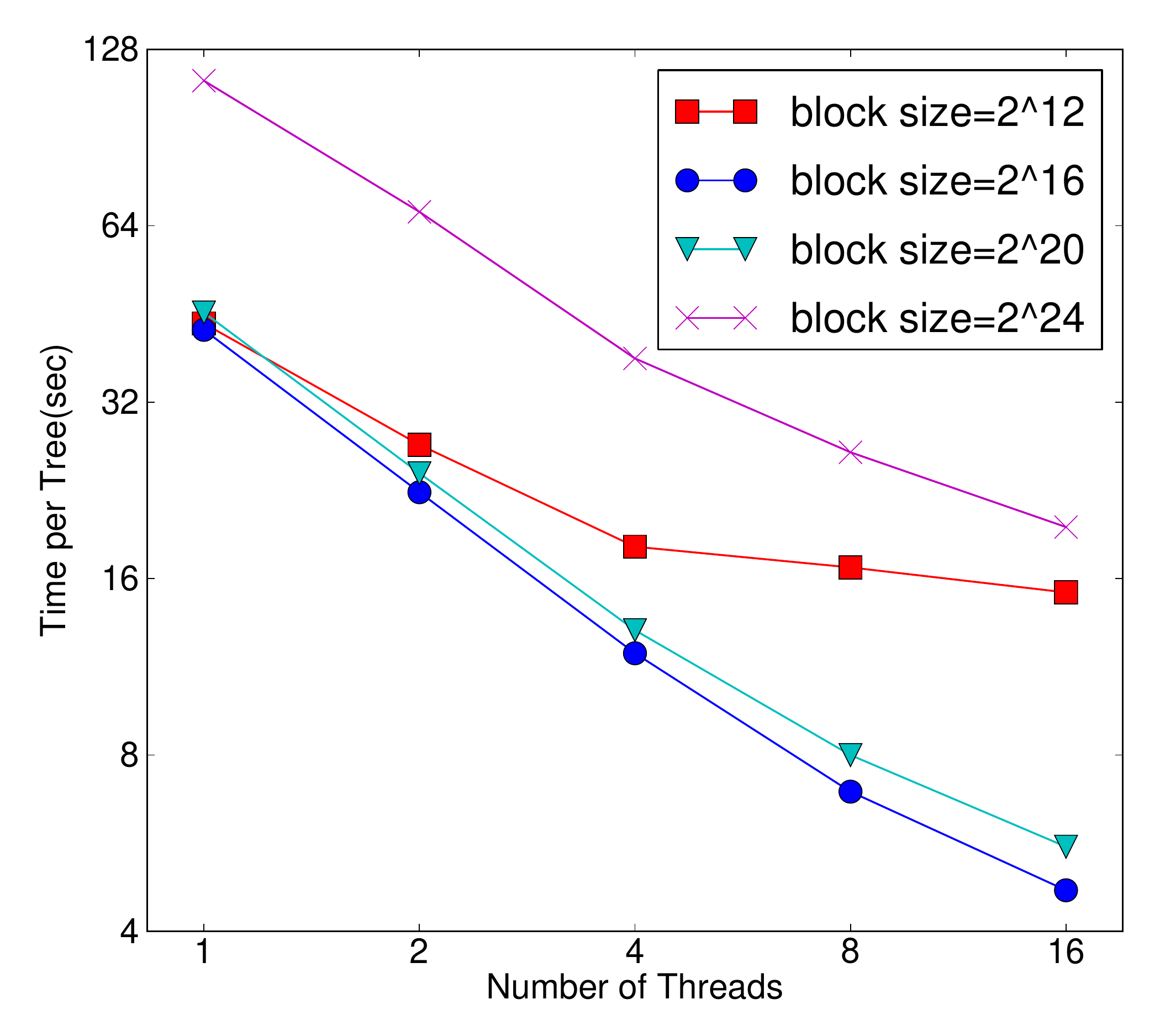}
  }
  \subfigure[Higgs 10M] {
    \includegraphics[width=.28\textwidth]{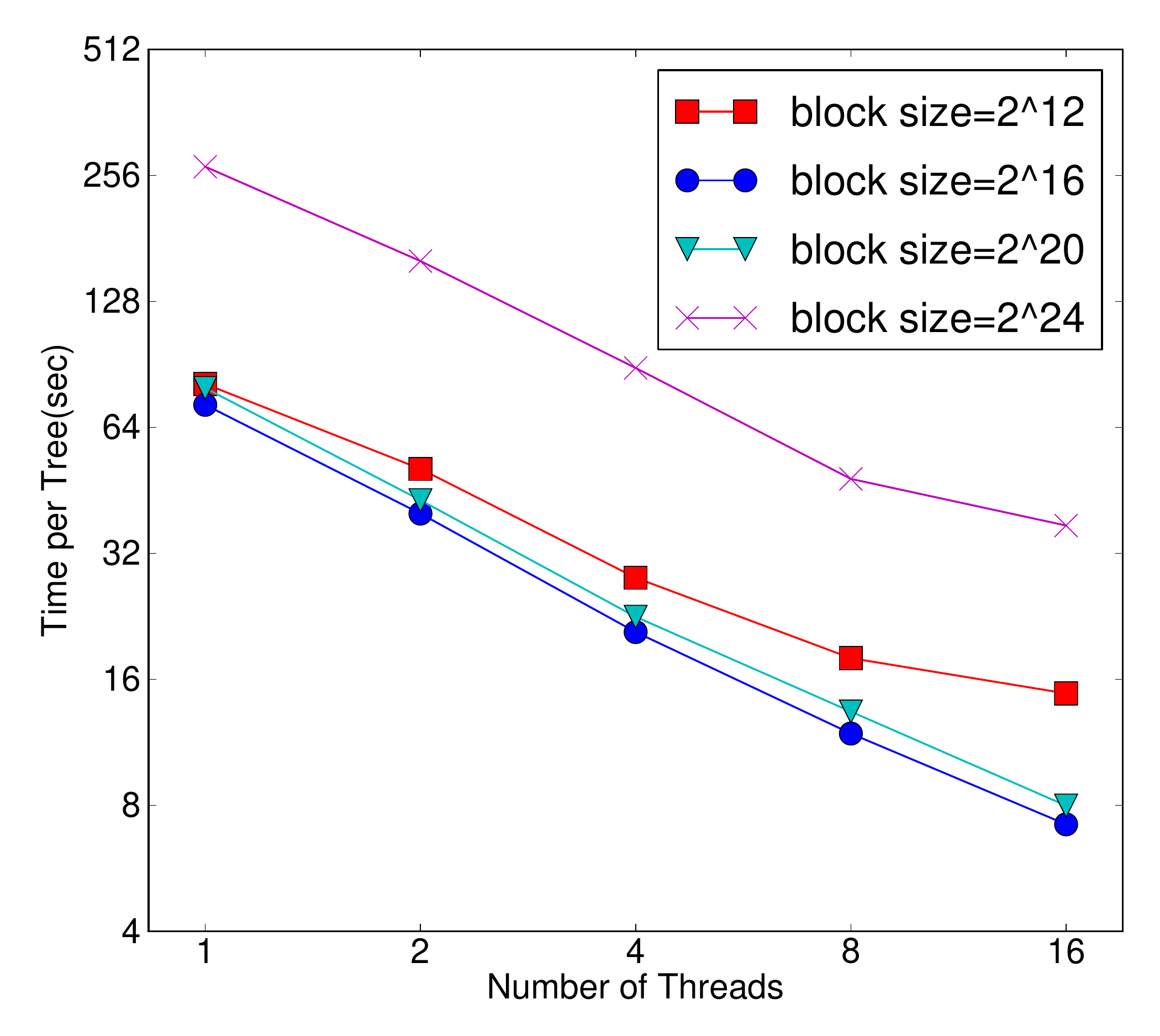}
  }
    \precap
  \caption{
    The impact of block size in the approximate algorithm.
    We find that overly small blocks results in inefficient parallelization, while overly large blocks also slows down training due to cache misses.
  } \label{fig:blocksize}
\postcap
\end{figure}
While the proposed block structure helps optimize the computation complexity of split finding, the new algorithm requires indirect fetches of gradient statistics by row index, since these values are accessed in order of feature. This is a non-continuous memory access.
A naive implementation of split enumeration introduces immediate read/write dependency between the accumulation and the non-continuous memory fetch operation~(see Fig.~\ref{fig:cache-miss}).
This slows down split finding when the gradient statistics do not fit into CPU cache and cache miss occur.

\begin{table*}
\centering
\caption{Comparison of major tree boosting systems.}\label{tbl:system-cmp}
\begin{tabular}{|l|m{1cm}|m{1.7cm}|m{1.7cm}|m{1.7cm}|m{1.7cm}|m{1.7cm}|}\hline
  System & exact greedy & approximate global & approximate local & out-of-core &sparsity aware& parallel \\ \hline
  \textbf{XGBoost} & yes & yes  & yes &yes &yes & yes\\\hline
  pGBRT & no & no & yes & no & no & yes\\\hline
  Spark MLLib & no & yes  & no  & no & partially & yes\\\hline
  H2O & no &  yes  & no   & no & partially & yes\\\hline
  scikit-learn & yes  & no  & no & no & no & no\\ \hline
  R GBM & yes & no & no & no& partially &no\\\hline
\end{tabular}
\end{table*}

For the exact greedy algorithm, we can alleviate the problem by a cache-aware prefetching algorithm.
Specifically, we allocate an internal buffer in each thread, fetch the gradient statistics into it, and then perform accumulation in a mini-batch manner.
This prefetching changes the direct read/write dependency to a longer dependency and helps to reduce the runtime overhead when number of rows in the is large.
Figure~\ref{fig:exact-cache} gives the comparison of cache-aware vs. non cache-aware algorithm on the the Higgs and the Allstate dataset.
We find that cache-aware implementation of the exact greedy algorithm runs twice as fast as the naive version when the dataset is large.

For approximate algorithms, we solve the problem by choosing a correct block size. We define the block size to be maximum number of examples in contained in a block, as this reflects the cache storage cost of gradient statistics.
Choosing an overly small block size results in small workload for each thread and leads to inefficient parallelization. On the other hand, overly large blocks result in cache misses, as the gradient statistics do not fit into the CPU cache.
A good choice of block size balances these two factors.
We compared various choices of block size on two data sets.
The results are given in Fig.~\ref{fig:blocksize}.
This result validates our discussion and shows that choosing $2^{16}$ examples per block balances the cache property and parallelization.

\subsection{Blocks for Out-of-core Computation}
One goal of our system is to fully utilize a machine's resources to achieve scalable learning.
Besides processors and memory, it is important to utilize disk space to handle data that does not fit into main memory. To enable out-of-core computation, we divide the data into multiple blocks and store each block on disk.
During computation, it is important to use an independent thread to pre-fetch the block into a main memory buffer, so computation can happen in concurrence with disk reading.
However, this does not entirely solve the problem since the disk reading takes most of the computation time.
It is important to reduce the overhead and increase the throughput of disk IO.
We mainly use two techniques to improve the out-of-core computation.

\noindent \textbf{Block Compression}
The first technique we use is block compression.
The block is compressed by columns, and decompressed on the fly by an independent thread when loading into main memory.
This helps to trade some of the computation in decompression with the disk reading cost.
We use a general purpose compression algorithm for compressing the features values.
For the row index, we substract the row index by the begining index of the block and use a 16bit integer to store each offset.
This requires $2^{16}$  examples per block, which is confirmed to be  a good setting.
In most of the dataset we tested, we achieve roughly a 26\% to 29\% compression ratio.

\noindent \textbf{Block Sharding}
The second technique is to shard the data onto multiple disks in an alternative manner.
A pre-fetcher thread is assigned to each disk and fetches the data into an in-memory buffer. The training thread then alternatively reads the data from each buffer. This helps to increase the throughput of disk reading when multiple disks are available.

\presec
\section{Related Works}\label{sec:rel}
\postsec

Our system implements gradient boosting~\cite{friedman2001greedy}, which performs additive optimization in functional space.
Gradient tree boosting has been successfully used in classification~\cite{friedman2000additive}, learning to rank~\cite{burges2010ranknet},
structured prediction~\cite{Chen:AISTATS2015} as well as other fields. XGBoost incorporates a regularized model to prevent overfitting. This this resembles previous work on regularized greedy forest~\cite{Tong:RGF}, but simplifies the objective and algorithm for parallelization.
Column sampling is a simple but effective technique borrowed from RandomForest~\cite{Breiman:RF}.
While sparsity-aware learning is essential in other types of models such as linear models~\cite{REF08a}, few works on tree learning have considered this topic in a principled way. The algorithm proposed in this paper is the first unified approach to handle all kinds of sparsity patterns.

There are several existing works on parallelizing tree learning~\cite{tyree2011parallel,PLANet}.
Most of these algorithms fall into the approximate framework described in this paper.
Notably, it is also possible to partition data by columns~\cite{Ye:GBDT} and apply the exact greedy algorithm. This is also supported in our framework, and the techniques such as cache-aware pre-fecthing can be used to benefit this type of algorithm.
While most existing works focus on the algorithmic aspect of parallelization,
our work improves in two unexplored system directions: out-of-core computation and cache-aware learning.
This gives us insights on how the system and the algorithm can be jointly optimized and provides an end-to-end system that can handle large scale problems
with very limited computing resources.
We also summarize the comparison between our system and existing opensource implementations in Table~\ref{tbl:system-cmp}.

Quantile summary~(without weights) is a classical problem in the database community~\cite{Greenwald:SIGMOID01,Zhang:SSDBM}.
However, the approximate tree boosting algorithm reveals a more general problem -- finding quantiles on weighted data.
To the best of our knowledge, the weighted quantile sketch proposed in this paper is the first method to solve this problem.
The weighted quantile summary is also not specific to the tree learning and can benefit other applications in data science and machine learning in the future.

\presec
\section{End to End Evaluations}\label{sec:exp}
\postsec

\subsection{System Implementation}
We implemented XGBoost as an open source package\footnote{\url{https://github.com/dmlc/xgboost}}.
The package is portable and reusable.
It supports various weighted classification and rank objective functions, as well as user defined objective function.
It is available in popular languages such as python, R, Julia and integrates naturally with language native data science pipelines such as scikit-learn.
The distributed version is built on top of the rabit library\footnote{https://github.com/dmlc/rabit} for allreduce.
The portability of XGBoost makes it available in many ecosystems, instead of only being tied to a specific platform.
The distributed XGBoost runs natively on Hadoop, MPI Sun Grid engine.
Recently, we also enable distributed XGBoost on jvm bigdata stacks such as Flink and Spark.
The distributed version has also been integrated into cloud platform
Tianchi\footnote{https://tianchi.aliyun.com} of Alibaba.
We believe that there will be more integrations in the future.

\subsection{Dataset and Setup}\label{subsec:dataset}

\begin{table}
\centering
\caption{Dataset used in the Experiments.}\label{tbl:dataset}
\begin{tabular}{|l|c|c|l|}\hline
  Dataset & $n$ & $m$  & Task  \\ \hline
  Allstate  & 10 M & 4227  & Insurance claim classification \\\hline
  Higgs Boson        & 10 M & 28    & Event classification \\ \hline
  Yahoo LTRC   & 473K & 700  &  Learning to Rank\\\hline
  Criteo       & 1.7 B  & 67   &   Click through rate prediction\\\hline
\end{tabular}
\end{table}
We used four datasets in our experiments. A summary of these datasets is given in Table~\ref{tbl:dataset}.
In some of the experiments, we use a randomly selected subset of the data either due to slow baselines or to demonstrate the performance of the algorithm with varying dataset size.
We use a suffix to denote the size in these cases. For example
Allstate-10K means a subset of the Allstate dataset with 10K instances.

The first dataset we use is the Allstate insurance claim dataset\footnote{https://www.kaggle.com/c/ClaimPredictionChallenge}.
The task is to predict the likelihood and cost of an insurance claim given different risk factors.
In the experiment, we simplified the task to only predict the likelihood of an insurance claim.
This dataset is used to evaluate the impact of sparsity-aware algorithm in Sec.~\ref{subsec:sparse}.
Most of the sparse features in this data come from one-hot encoding.
We randomly select 10M instances as training set and use the rest as evaluation set.

The second dataset is the Higgs boson dataset\footnote{https://archive.ics.uci.edu/ml/datasets/HIGGS} from high energy physics.
The data was produced using Monte Carlo simulations of physics events.
It contains 21 kinematic properties measured by the particle detectors in the accelerator.
It also contains seven additional derived physics quantities of the particles.
The task is to classify whether an event corresponds to the Higgs boson.
We randomly select 10M instances as training set and use the rest as evaluation set.

The third dataset is the Yahoo! learning to rank challenge dataset~\cite{YahooLTRC}, which is one of the most commonly used benchmarks in learning to rank algorithms.
The dataset contains 20K web search queries, with each query corresponding to a list of around 22 documents.
The task is to rank the documents according to relevance of the query.
We use the official train test split in our experiment.

The last dataset is the criteo terabyte click log dataset\footnote{http://labs.criteo.com/downloads/download-terabyte-click-logs/}.
We use this dataset to evaluate the scaling property of the system in the out-of-core and the distributed settings.
The data contains 13 integer features and 26 ID features of user, item and advertiser information.
Since a tree based model is better at handling continuous features,
we preprocess the data by calculating the statistics of average CTR and count of ID features on the first ten days,  replacing the ID features by the corresponding count statistics during the next ten days for training.
The training set after preprocessing contains 1.7 billion instances with 67 features~(13 integer, 26 average CTR statistics and 26 counts).
The entire dataset is more than one terabyte in LibSVM format.

We use the first three datasets for the single machine parallel setting, and the last dataset for the distributed and out-of-core settings.
All the single machine experiments are conducted on a Dell PowerEdge R420 with two eight-core Intel Xeon (E5-2470) (2.3GHz) and 64GB of memory.
If not specified, all the experiments are run using all the available cores in the machine.
The machine settings of the distributed and the out-of-core experiments will be described in the corresponding section.
In all the experiments, we boost trees with a common setting of maximum depth equals 8, shrinkage equals 0.1
and no column subsampling unless explicitly specified.
We can find similar results when we use other settings of maximum depth.

\subsection{Classification}
\begin{table}[t!]
\centering
\caption{Comparison of Exact Greedy Methods with 500 trees on Higgs-1M data.}\label{tbl:higgs-cmp}
\begin{tabular}{|l|c|c|}\hline
  Method & Time per Tree (sec) & Test AUC   \\ \hline
  XGBoost  &  0.6841 &  0.8304 \\ \hline
  XGBoost (colsample=0.5) & 0.6401 & 0.8245 \\ \hline
  scikit-learn  & 28.51  & 0.8302  \\ \hline
  R.gbm    &  1.032 &  0.6224  \\ \hline
\end{tabular}
\end{table}

In this section, we evaluate the performance of XGBoost on a single machine using the exact greedy algorithm on Higgs-1M data, by comparing it against two other commonly used exact greedy tree boosting implementations.
Since scikit-learn only handles non-sparse input, we choose the dense Higgs dataset for a fair comparison.
We use the 1M subset to make scikit-learn finish running in reasonable time.
Among the methods in comparison, R's GBM uses a greedy approach that only expands one branch of a tree, which makes
it faster but can result in lower accuracy, while both scikit-learn and XGBoost learn a full tree.
The results are shown in Table~\ref{tbl:higgs-cmp}.
Both XGBoost and scikit-learn give better performance than R's GBM, while XGBoost runs more than 10x faster than scikit-learn.
In this experiment, we also find column subsamples gives slightly worse performance than using all the features. This could due to the fact that there are few important features in this dataset and we can benefit from greedily select from all the features.

\subsection{Learning to Rank}
\begin{figure}[t]
  \centering
  \includegraphics[width=.35\textwidth]{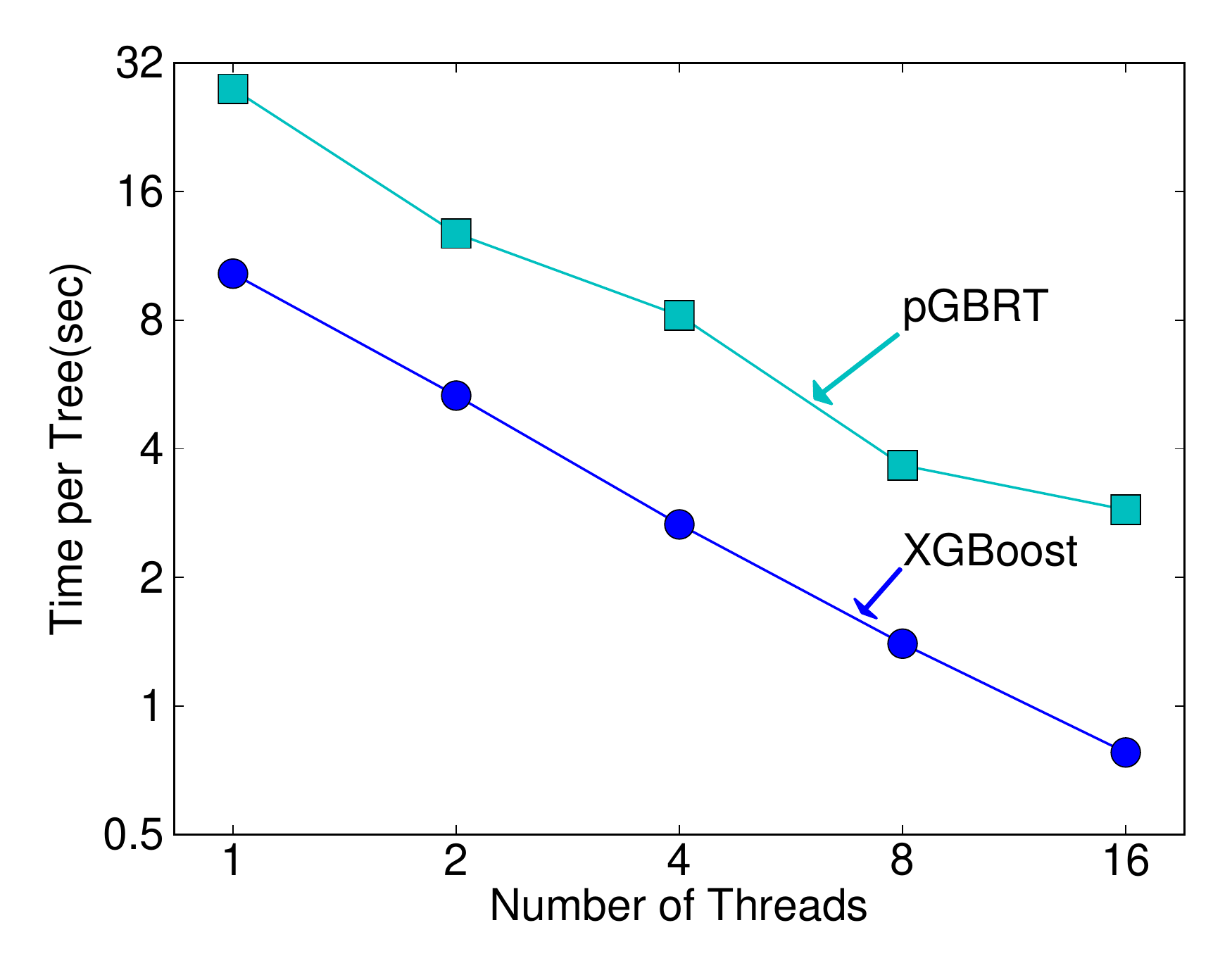}
    \precap
  \caption{Comparison between XGBoost and pGBRT on Yahoo LTRC dataset.
  }
\postcap
\label{fig:sparse-yahoo}
\end{figure}
\begin{table}[t!]
\centering
\caption{Comparison of Learning to Rank with 500 trees on Yahoo! LTRC Dataset}\label{tbl:yahoo-cmp}
\begin{tabular}{| l |c|c|}\hline
  Method & Time per Tree (sec) & NDCG@10   \\ \hline
  XGBoost & 0.826 &  0.7892  \\ \hline
  XGBoost (colsample=0.5)  & 0.506 & 0.7913  \\ \hline
  pGBRT~\cite{tyree2011parallel} & 2.576 &  0.7915  \\ \hline
\end{tabular}
\end{table}

\begin{figure}[!t]
  \centering
  \includegraphics[width=.36\textwidth]{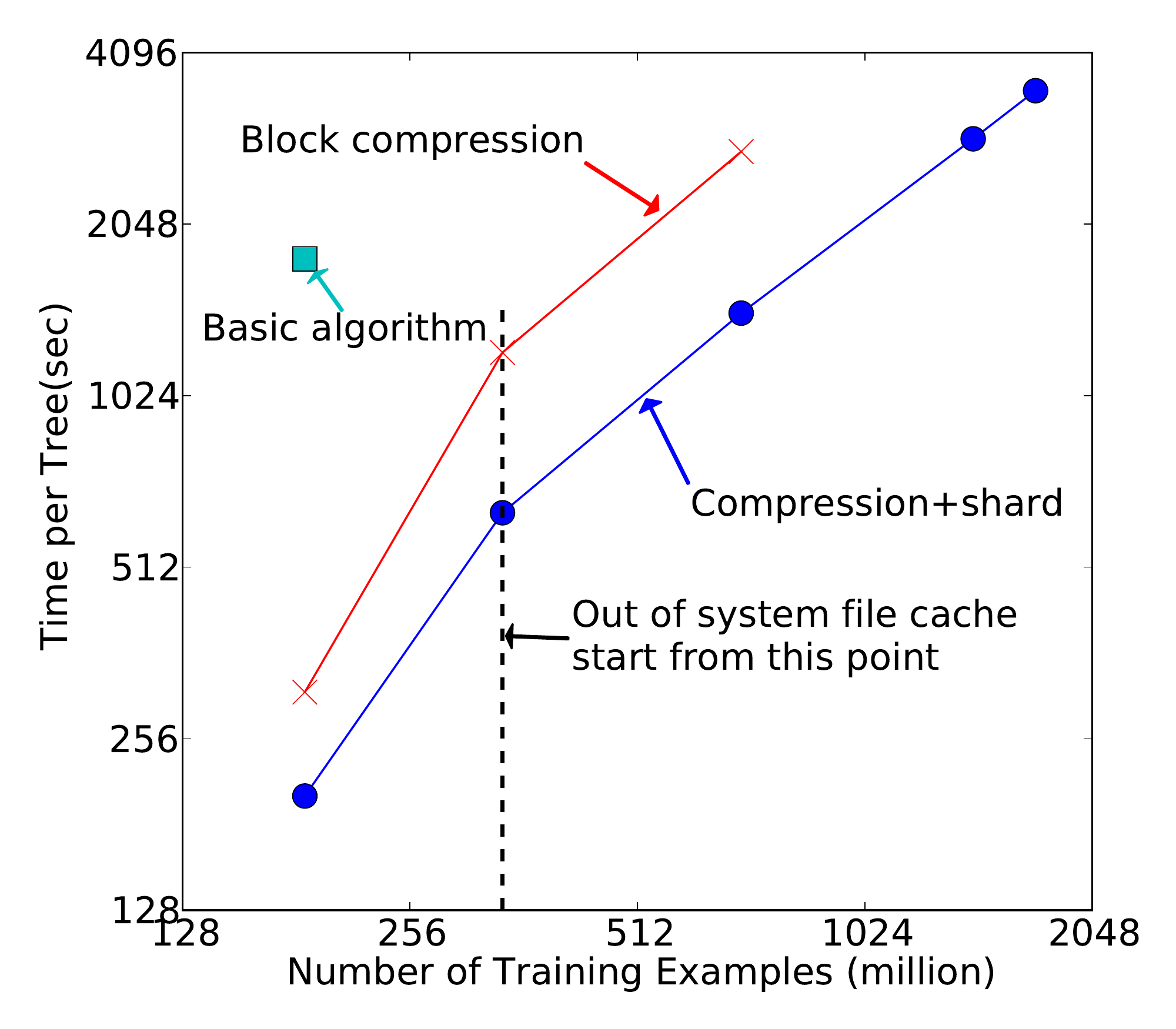}
    \precap
  \caption{Comparison of out-of-core methods on different subsets of criteo data.
    The missing data points are due to out of disk space.
    We can find that basic algorithm can only handle 200M examples.
    Adding compression gives 3x speedup, and sharding into two disks gives another 2x speedup.
    The system runs out of file cache start from 400M examples. The algorithm really has to rely on disk after this point.
    The compression+shard method has a less dramatic slowdown when running out of file cache, and exhibits a linear trend afterwards.
  }\label{fig:disk}
    \postcap
\end{figure}

\begin{figure}[t!]
  \centering
  \subfigure[End-to-end time cost include data loading] {
    \includegraphics[width=.37\textwidth]{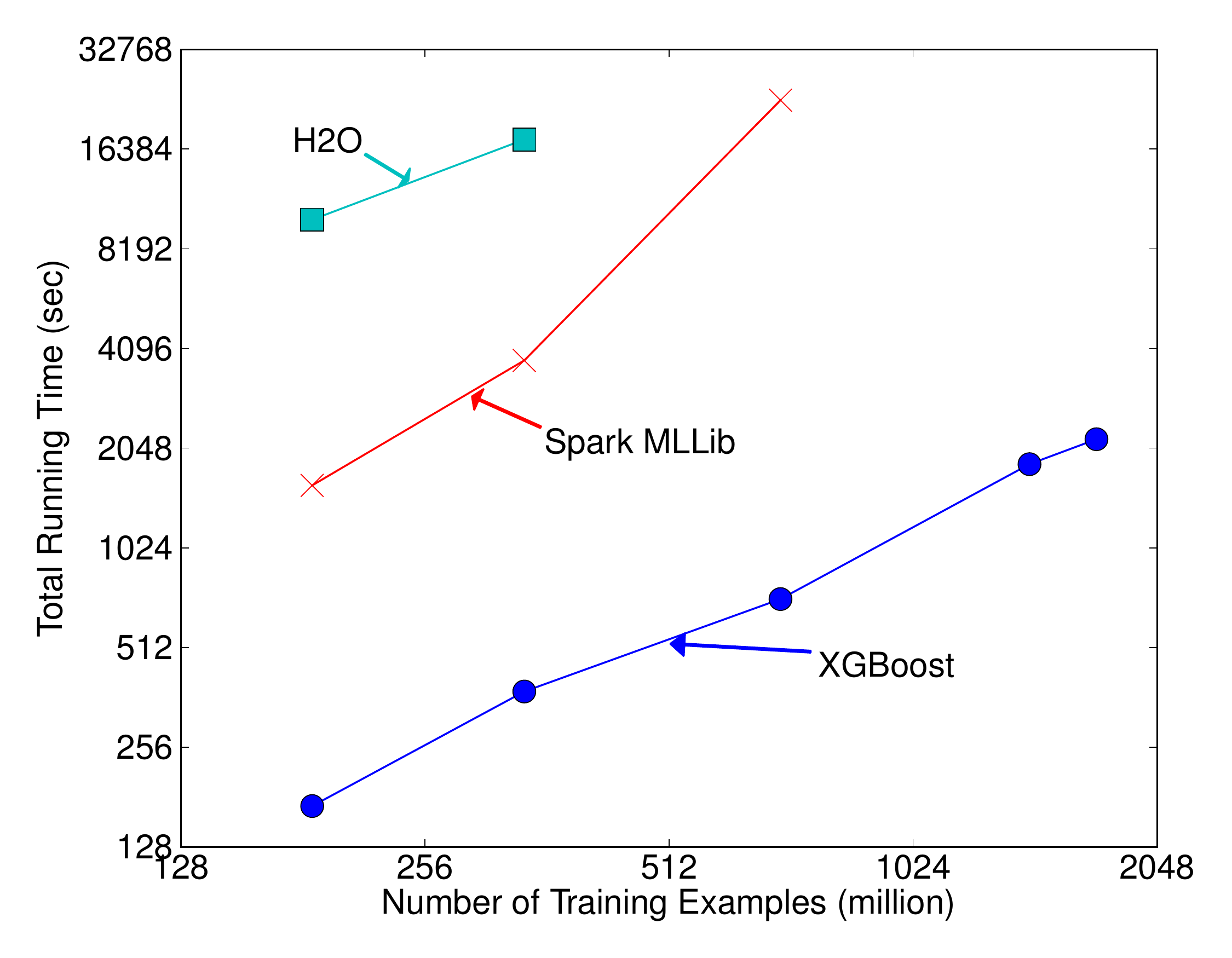}
  }
  \subfigure[Per iteration cost exclude data loading] {
    \includegraphics[width=.37\textwidth]{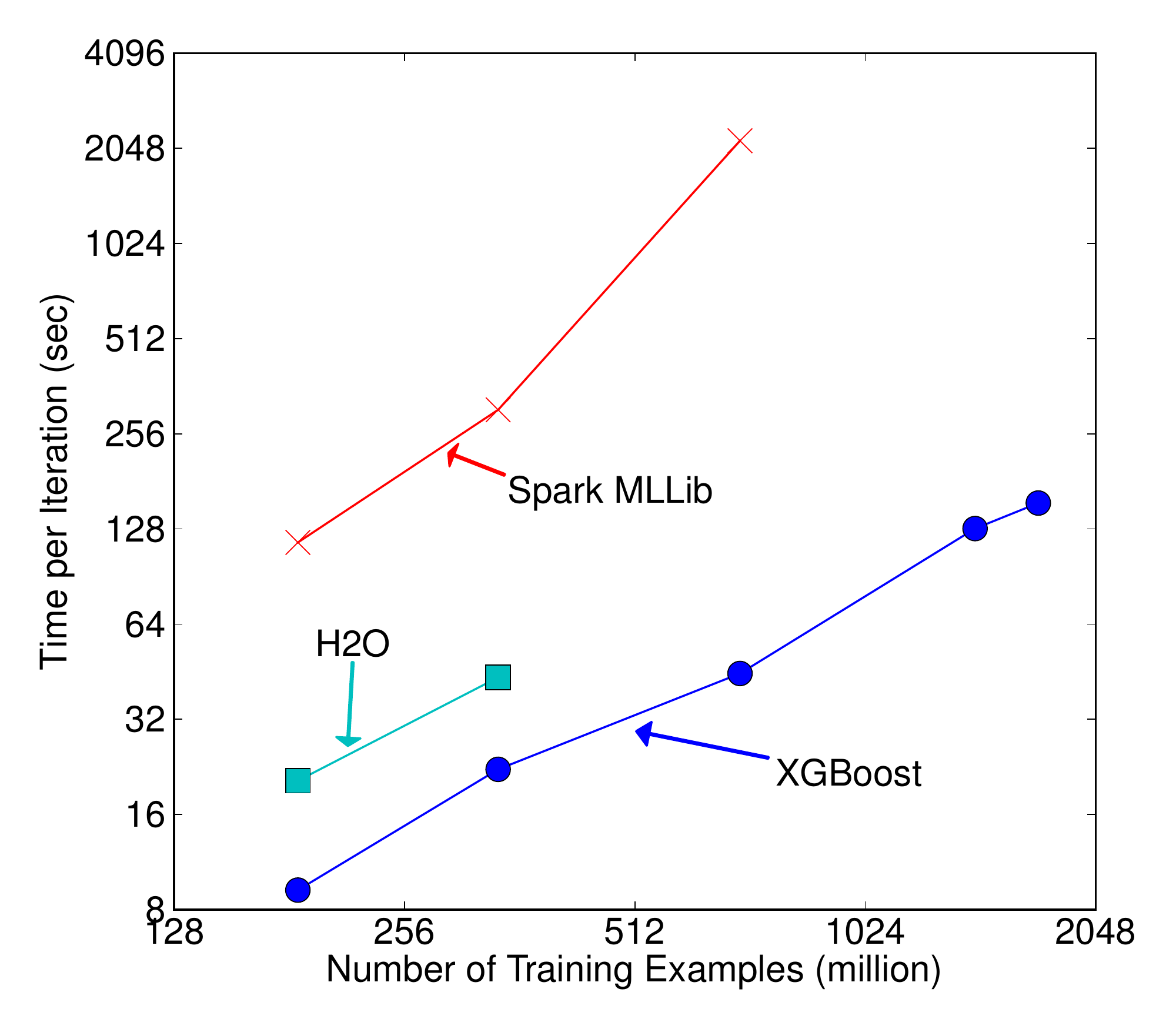}
  }
  \precap
  \caption{
    Comparison of different distributed systems on 32 EC2 nodes for 10 iterations on different subset of criteo data.
    XGBoost runs more 10x than spark per iteration and 2.2x as H2O's optimized version
    ~(However, H2O is slow in loading the data, getting worse end-to-end time).
    Note that spark suffers from drastic slow down when running out of memory.
    XGBoost runs faster and scales smoothly to the full 1.7 billion examples with given resources by utilizing out-of-core computation.
  }\label{fig:bigdata}
  \postcap
\end{figure}

\begin{figure}[t!]
  \centering
   \includegraphics[width=.35\textwidth]{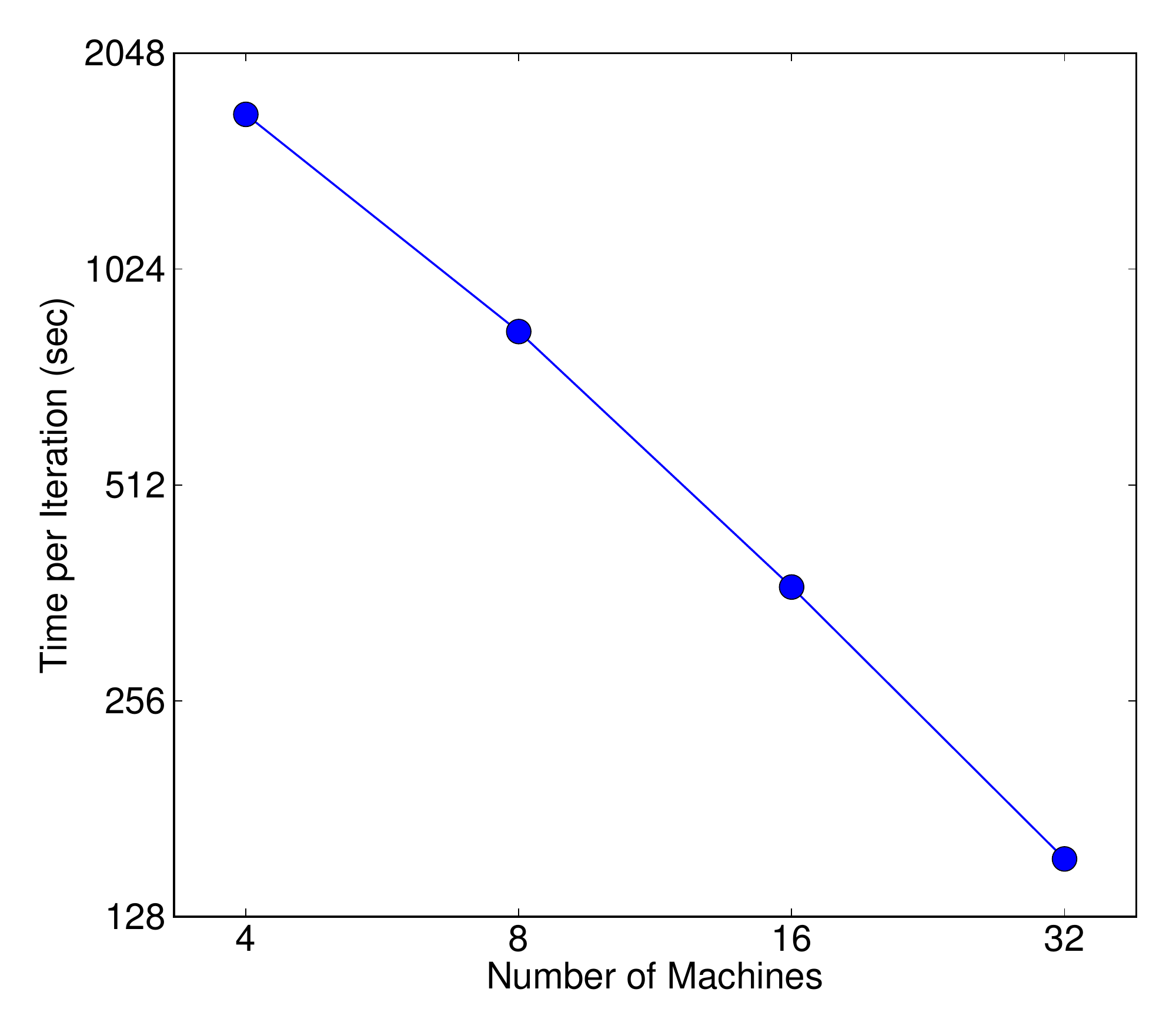}
    \precap
  \caption{
    Scaling of XGBoost with different number of machines on criteo full 1.7 billion dataset.
    Using more machines  results in more file cache and makes the system run faster,
    causing the trend to be slightly super linear.
    XGBoost can process the entire dataset using as little as four machines, and scales smoothly by utilizing more available resources.
  }\label{fig:scale}
    \postcap
\end{figure}

We next evaluate the performance of XGBoost on the learning to rank problem.
We compare against pGBRT~\cite{tyree2011parallel}, the best previously pubished system on this task.
XGBoost runs exact greedy algorithm, while pGBRT only support an approximate algorithm.
The results are shown in Table~\ref{tbl:yahoo-cmp} and Fig.~\ref{fig:sparse-yahoo}.
We find that XGBoost runs faster.
Interestingly, subsampling columns not only reduces running time, and but also gives a bit higher performance for this problem.
This could due to the fact that the subsampling helps prevent overfitting, which is observed by many of the users.

\subsection{Out-of-core Experiment}
We also evaluate our system in the out-of-core setting on the criteo data.
We conducted the experiment on one AWS c3.8xlarge machine~(32 vcores, two 320 GB SSD, 60 GB RAM).
The results are shown in Figure~\ref{fig:disk}.
We can find that compression helps to speed up computation by factor of three, and sharding into two disks further gives 2x speedup.
For this type of experiment, it is important to use a very large dataset to drain the system file cache for a real out-of-core setting.
This is indeed our setup. We can observe a transition point when the system runs out of file cache.
Note that the transition in the final method is less dramatic. This is due to larger disk throughput and better utilization of computation resources.
Our final method is able to process 1.7 billion examples on a single machine.

\subsection{Distributed Experiment}

Finally, we evaluate the system in the distributed setting.  We set up a YARN cluster on EC2 with  m3.2xlarge machines, which is a very common choice for clusters.
Each machine contains 8 virtual cores, 30GB of RAM and two 80GB SSD local disks. The dataset is stored on AWS S3 instead of HDFS to avoid purchasing persistent storage.

We first compare our system against two production-level distributed systems: Spark MLLib~\cite{MLLib} and H2O~\footnote{www.h2o.ai}.
We use 32 m3.2xlarge machines and test the performance of the systems with various input size.
Both of the baseline systems are in-memory analytics frameworks that need to store the data in RAM, while XGBoost can switch to out-of-core setting when it runs out of memory. The results are shown in Fig.~\ref{fig:bigdata}.
We can find that XGBoost runs faster than the baseline systems.
More importantly, it is able to take advantage of out-of-core computing and smoothly scale to all 1.7 billion examples with the given limited computing resources.
The baseline systems are only able to handle subset of the data with the given resources.
This experiment shows the advantage to bring all the system improvement together and solve a real-world scale problem.
We also evaluate the scaling property of XGBoost by varying the number of machines. The results are shown in Fig.~\ref{fig:scale}.
We can find XGBoost's performance scales linearly as we add more machines. Importantly, XGBoost is able to handle the entire 1.7 billion
data with only four machines. This shows the system's potential to handle even larger data.
\presec
\section{Conclusion}\label{sec:con}
\postsec
In this paper, we described the lessons we learnt when building XGBoost, a scalable tree boosting system that is widely used by data scientists and provides state-of-the-art results on many problems.
We proposed a novel sparsity aware algorithm for handling sparse data and a theoretically justified weighted quantile sketch for approximate learning.  Our experience shows that cache access patterns, data compression and sharding are essential elements for building a scalable end-to-end system for tree boosting. These lessons can be applied to other machine learning systems as well.
By combining these insights, XGBoost is able to solve real-world scale problems using a minimal amount of resources.

%ACKNOWLEDGMENTS are optional
\presec
\section*{Acknowledgments}
\postsec
\small{
We would like to thank Tyler B. Johnson, Marco Tulio Ribeiro, Sameer Singh, Arvind Krishnamurthy for their valuable feedback.
We also sincerely thank Tong He, Bing Xu, Michael Benesty, Yuan Tang, Hongliang Liu, Qiang Kou, Nan Zhu and all other contributors in the XGBoost community.
This work was supported in part by ONR (PECASE) N000141010672, NSF IIS 1258741 and the TerraSwarm Research
Center sponsored by MARCO and DARPA.
}

\bibliographystyle{abbrv}
\small{
\bibliography{XGBoost}
}  % sigproc.bib is the name of the Bibliography in this case

\appendix

\section{Weighted Quantile Sketch}
In this section, we introduce the weighted quantile sketch algorithm.
Approximate answer of quantile queries is for many real-world applications.
One classical approach to this problem is GK algorithm~\cite{Greenwald:SIGMOID01} and extensions based on the GK framework~\cite{Zhang:SSDBM}.
The main component of these algorithms is a data structure called quantile summary, that is able to answer quantile queries with relative accuracy of $\eps$.
Two operations are defined for a quantile summary:
\begin{itemize}
\item A merge operation that combines two summaries with approximation error $\eps_1$ and $\eps_2$ together and create a merged summary with approximation error $\max(\eps_1,\eps_2)$.
\item A prune operation that reduces the number of elements in the summary to $b+1$ and changes approximation error from $\eps$ to $\eps + \frac{1}{b}$.
\end{itemize}
A quantile summary with merge and prune operations forms basic building blocks of the  distributed and streaming quantile computing algorithms~\cite{Zhang:SSDBM}.

In order to use quantile computation for approximate tree boosting, we need to find quantiles on weighted data. This more general problem is not supported
by any of the existing algorithm.  In this section, we describe a non-trivial weighted quantile summary structure to solve this problem.
Importantly, the new algorithm contains merge and prune operations with \emph{the same guarantee} as GK summary.
This allows our summary to be plugged into all the frameworks used GK summary as building block and answer quantile queries over weighted data efficiently.

\subsection{Formalization and Definitions}
Given an input multi-set $\sD=\{(x_1,w_1), (x_2,w_2) \cdots (x_n,w_n)\}$  such that $w_i\in \Rplus, x_i \in \sX$.
Each $x_i$ corresponds to a position of the point and $w_i$ is the weight of the point. Assume we have a total order $<$ defined on $\sX$.
 Let us define  two rank functions $r^-_{\sD}, r^+_{\sD}: \sX \rightarrow \Rplus$
\begin{equation}
    r^-_{\sD}(y) = \sum_{(x,w)\in \sD, x < y} w
\end{equation}
\begin{equation}
    r^+_{\sD}(y) = \sum_{(x,w)\in \sD, x \leq y} w
\end{equation}
We should note that since $\sD$ is defined to be a \emph{multiset} of the points. It can contain multiple record with exactly same position $x$ and weight $w$.
We also define another weight function $\w_{\sD}:  \sX \rightarrow \Rplus$ as
\begin{equation}
    \w_{\sD}(y) =  r^+_{\sD}(y)  - r^-_{\sD}(y) =  \sum_{(x,w)\in \sD, x = y} w .
\end{equation}
Finally, we also define the weight of multi-set $\sD$ to be  the sum of weights of all the points in the set
\begin{equation}
    \w(\sD) = \sum_{(x,w)\in \sD} w
\end{equation}
Our task is given a series of input $\sD$, to estimate $r^+(y)$ and $r^-(y)$ for $y\in \sX$ as well as finding points
with specific rank.
Given these notations, we define quantile summary of weighted examples as follows:

\begin{thm:def}\label{def:sketch}{Quantile Summary of Weighted Data}\\
A quantile summary for $\sD$ is defined to be tuple $Q(\sD) = (S, \tdr^+, \tdr^-, \tdw)$, where
$S = \{x_1, x_2,\cdots, x_k\}$ is selected from the points in $\sD$~(i.e. $x_i \in\{x|(x,w)\in \sD\}$) with the following properties:

1) $x_{i}< x_{i+1} \mbox{ for all } i$, and $x_1$ and $x_k$ are minimum and maximum point in $\sD$:
$$x_1 = \min_{(x,w)\in \sD} x,\ \  x_k = \max_{(x,w)\in \sD} x$$
2) $\tdr^+$, $\tdr^-$ and $\tdw$ are functions in $S\rightarrow \Rplus$, that satisfies
\begin{equation}\label{eq:constraint}
     \tdr^-(x_i) \leq r^-_{\sD}(x_i),\ \  \tdr^+(x_i) \geq r^+_{\sD}(x_i), \ \ \tdw(x_i)\leq \w_{\sD}(x_i),
\end{equation}
 the equality sign holds for maximum and minimum point~(
$ \tdr^-(x_i) = r^-_{\sD}(x_i)$, $\tdr^+(x_i) = r^+_{\sD}(x_i)$ and $\tdw(x_i)= \w_{\sD}(x_i) $ for $i\in \{1, k\}$). \\
Finally, the function value must also satisfy the following constraints
\begin{equation}
 \tdr^-(x_i) + \tdw(x_i)\leq \tdr^-(x_{i+1}), \ \ \tdr^+(x_i) \leq \tdr^+(x_{i+1}) - \tdw(x_{i+1})
\end{equation}
\end{thm:def}

Since these functions are only defined on $S$, it is suffice to use $4 k$ record to store the summary. Specifically, we need to remember each $x_i$ and the corresponding function values of each $x_i$.

\begin{thm:def}\label{def:extend}{Extension of Function Domains}\\
Given a quantile summary $Q(\sD) = (S, \tdr^+, \tdr^-, \tdw)$ defined in Definition~\ref{def:sketch}, the domain of $\tdr^+$, $\tdr^-$ and $\tdw$ were defined only in $S$. We extend the definition of these functions to $\sX \rightarrow \Rplus$ as follows \\
\noindent When $y < x_1$:
\begin{equation}
    \tdr^-(y) = 0, \  \tdr^+(y) = 0,\   \tdw(y) = 0
\end{equation}
\noindent When $y > x_k$:
\begin{equation}
    \tdr^-(y) = \tdr^+(x_k), \  \tdr^+(y) = \tdr^+(x_k),\   \tdw(y) = 0
\end{equation}
\noindent When $y \in (x_i, x_{i+1})$ for some $i$:
\begin{equation}
\begin{split}
    \tdr^-(y)& =\tdr^-(x_i)+\tdw(x_i), \\
      \tdr^+(y)& = \tdr^+(x_{i+1}) - \tdw(x_{i+1}),\\   \tdw(y)& = 0
\end{split}
\end{equation}
\end{thm:def}

\begin{thm:lemma}\label{lem:ext-constraint}{Extended Constraint}\\
The extended definition of $\tdr^-$, $\tdr^+$, $\tdw$ satisfies the following constraints
\begin{equation}\label{eq:extend-1}
     \tdr^-(y) \leq r^-_{\sD}(y),\ \  \tdr^+(y) \geq r^+_{\sD}(y), \ \ \tdw(y)\leq \w_{\sD}(y)
\end{equation}
\begin{equation} \label{eq:extend-2}
\tdr^-(y) + \tdw(y)\leq \tdr^-(x), \ \ \tdr^+(y) \leq \tdr^+(x) - \tdw(x), \mbox{ for all } y < x
\end{equation}
\end{thm:lemma}
\begin{proof}
The only non-trivial part is to prove the case when $y \in (x_i, x_{i+1})$:
$$
\tdr^-(y) = \tdr^-(x_i)+\tdw(x_i) \leq r_{\sD}^-(x_i) +   \w_{\sD}(x_i) \leq  r_{\sD}^-(y)
$$
$$
\tdr^+(y) = \tdr^+(x_{i+1})-\tdw(x_{i+1}) \geq r_{\sD}^+(x_{i+1}) - \w_{\sD}(x_{i+1}) \geq  r_{\sD}^+(y)
$$
This proves Eq.~\eqref{eq:extend-1}.
Furthermore, we can verify that
$$
\tdr^+(x_{i}) \leq \tdr^+(x_{i+1}) - \tdw(x_{i+1}) =  \tdr^+(y) -\tdw(y)
$$
$$
\tdr^-(y) +\tdw(y) = \tdr^-(x_i)+\tdw(x_i) + 0 \leq   \tdr^-(x_{i+1})
$$
$$
\tdr^+(y)= \tdr^+(x_{i+1}) - \tdw(x_{i+1})
$$
Using these facts and transitivity of $<$ relation, we can prove Eq.~\eqref{eq:extend-2}
\end{proof}
We should note that the extension is based on the ground case defined in $S$, and we do not require extra space to store the summary in order to use the extended definition. We are now ready to introduce the definition of $\eps$-approximate quantile summary.
\begin{thm:def}{$\eps$-Approximate Quantile Summary}\\
Given a quantile summary $Q(\sD) = (S, \tdr^+, \tdr^-, \tdw)$, we call it is $\eps$-approximate summary if for any $ y \in \sX$
\begin{equation}\label{eq:epsdef}
\tdr^{+}(y) - \tdr^{-}(y) - \tdw(y) \leq  \eps \w(\sD)
\end{equation}
We use this definition since we know that $r^-(y)\in[\tdr^{-}(y), \tdr^{+}(y) -\tdw(y)] $ and $r^+(y)\in[\tdr^{-}(y)+\tdw(y), \tdr^{+}(y)] $. Eq.~\eqref{eq:epsdef} means the we can get estimation of $r^+(y)$ and $r^-(y)$ by error of at most $ \eps \w(\sD)$.
\end{thm:def}
\begin{thm:lemma}\label{lemma:eps}
Quantile summary $Q(\sD) = (S, \tdr^+, \tdr^-, \tdw)$ is an $\eps$-approximate summary if and only if the following two condition holds
\begin{equation}\label{eq:constraint22}
\tdr^{+}(x_i) - \tdr^{-}(x_i) - \tdw(x_i) \leq  \eps \w(\sD)
\end{equation}
\begin{equation}\label{eq:constraint2}
\tdr^{+}(x_{i+1}) - \tdr^{-}(x_i) - \tdw(x_{i+1}) - \tdw(x_{i}) \leq  \eps \w(\sD)
\end{equation}
\begin{proof}
The key is again consider $y\in(x_i, x_{i+1})$
$$
\tdr^{+}(y) - \tdr^{-}(y) - \tdw(y) =  [\tdr^+(x_{i+1})-\tdw(x_{i+1})]  - [ \tdr^+(x_{i})+\tdw(x_{i})] - 0
$$
This means the condition in Eq.~\eqref{eq:constraint2} plus Eq.\eqref{eq:constraint22} can give us Eq.~\eqref{eq:epsdef}
\end{proof}
\end{thm:lemma}

\noindent \textbf{Property of Extended Function} In this section, we have introduced the extension of function $\tdr^+, \tdr^-, \tdw$ to $\sX\rightarrow \Rplus$.
The key theme discussed in this section is the relation of \emph{constraints on the original function and constraints on the extended function}.
Lemma~\ref{lem:ext-constraint} and ~\ref{lemma:eps} show that the constraints on the original function can lead to in more general constraints on the extended function.
This is a very useful property which will be used in the proofs in later sections.

\subsection{Construction of Initial Summary}
Given a small multi-set $\sD = \{(x_1,w_1), (x_2,w_2),\cdots, (x_n,w_n)\}$, we can construct initial summary $Q(\sD) = \{S, \tdr^+, \tdr^-, \tdw\}$, with $S$ to the set of all values in $\sD$~($S = \{x|(x,w)\in \sD\}$), and $\tdr^+, \tdr^-, \tdw$ defined to be
\begin{equation}
\tdr^{+}(x)  = r^+_{\sD}(x), \ \  \tdr^{-}(x)  = r^-_{\sD}(x), \ \ \tdw(x) = \w_{\sD}(x) \mbox{ for } x \in S
\end{equation}
The constructed summary is $0$-approximate summary, since it can answer all the queries accurately. The constructed summary can be feed into future operations described in the latter sections.

\subsection{Merge Operation}
In this section, we define how we can merge the two summaries together. Assume we have $Q(\sD_1) =(S_1, \tdrx{1}^+, \tdrx{1}^-, \tdwx{1})$ and  $Q(\sD_2) =(S_2, \tdrx{1}^+, \tdrx{2}^-, \tdwx{2})$ quantile summary of two dataset $\sD_1$ and $\sD_2$. Let $\sD = \sD_1 \cup \sD_2$, and define the merged summary $Q(\sD)=(S, \tdr^+, \tdr^-, \tdw)$ as follows.
\begin{equation}
S=\{x_1, x_2\cdots, x_k\}, x_i \in S_1 \mbox{ or } x_i \in S_2
\end{equation}
The points in $S$ are combination of points in $S_1$ and $S_2$. And the function $\tdr^+, \tdr^-,\tdw$ are defined to be
\begin{equation}
\tdr^-(x_i) = \tdrx{1}^-(x_i) + \tdrx{2}^-(x_i)
\end{equation}
\begin{equation}
\tdr^+(x_i) = \tdrx{1}^+(x_i) + \tdrx{2}^+(x_i)
\end{equation}
\begin{equation}
\tdw(x_i) =\tdwx{1}(x_i)+ \tdwx{2}(x_{i})
\end{equation}
Here we use functions defined on $S\rightarrow \Rplus$ on the left sides of equalities
and use the  extended function definitions on the right sides.

Due to additive nature of $r^+$, $r^-$ and $\w$, which can be formally written as
\begin{equation}
\begin{split}
r^-_{\sD}(y) =&r^-_{\sD_1}(y)+ r^-_{\sD_2}(y),\\
 r^+_{\sD}(y) =&r^+_{\sD_1}(y)+ r^+_{\sD_2}(y),\\
  \w_{\sD}(y) =&\w_{\sD_1}(y)+ \w_{\sD_2}(y),
\end{split}
\end{equation}
and the extended constraint property in Lemma~\ref{lem:ext-constraint}, we can verify that
$Q(\sD)$ satisfies all the constraints in Definition~\ref{def:sketch}. Therefore it is a valid quantile summary.

\begin{thm:lemma}\label{lem:merge-extend}
The combined quantile summary satisfies
\begin{equation}
\tdr^-(y) = \tdrx{1}^-(y) + \tdrx{2}^-(y)
\end{equation}
\begin{equation}
\tdr^+(y) = \tdrx{1}^+(y) + \tdrx{2}^+(y)
\end{equation}
\begin{equation}
\tdw(y) =\tdwx{1}(y)+ \tdwx{2}(y)
\end{equation}
for all $y\in \sX$
\end{thm:lemma}
This can be obtained by straight-forward application of Definition~\ref{def:extend}.
\begin{thm:thm}
If $Q(\sD_1)$ is $\eps_1$-approximate summary, and  $Q(\sD_2)$ is $\eps_2$-approximate summary. Then the merged summary $Q(\sD)$ is $\max(\eps_1,\eps_2)$-approximate summary.
\end{thm:thm}
\begin{proof}
For any $y\in \sX$, we have
\begin{equation*}
\begin{split}
&\tdr^+(y) - \tdr^-(y) -\tdw(y)\\
=& [\tdrx{1}^+(y) +\tdrx{2}^+(y)] - [\tdrx{1}^-(y)+\tdrx{2}^-(y)] -[\tdwx{1}(y) + \tdwx{2}(y)]\\
\leq& \eps_1 \w(\sD_1)+\eps_2\w(\sD_2) \leq \max(\eps_1,\eps_2) \w(\sD_1\cup \sD_2)
\end{split}
\end{equation*}
Here the first inequality is due to Lemma~\ref{lem:merge-extend}.
\end{proof}

\subsection{Prune Operation}
\begin{algorithm}[t]
\caption{Query Function $g(Q,d)$}\label{alg:query}
\KwIn{$d$: $0 \leq d \leq \w(\sD)$}
\KwIn{$Q(\sD) = (S, \tdr^+, \tdr^-, \tdw)$ where $S={x_1,x_2,\cdots,x_k}$}
\lIf{$d < \frac{1}{2}[\tdr^-(x_1) + \tdr^+(x_1) ]$}{
    \Return $x_1$
}
\lIf{$d \geq \frac{1}{2}[\tdr^-(x_k) + \tdr^+(x_k) ]$}{
    \Return $x_k$
}
Find $i$ such that $\frac{1}{2}[\tdr^-(x_i) + \tdr^+(x_i) ] \leq d < \frac{1}{2}[\tdr^-(x_{i+1}) + \tdr^+(x_{i+1}) ]$\\
\uIf{$ 2 d  < \tdr^-(x_i) + \tdw(x_i) + \tdr^+(x_{i+1}) - \tdw(x_{i+1}) $} {
    \Return $x_i$
}\Else{
    \Return $x_{i+1}$
}
\end{algorithm}
Before we start discussing the prune operation, we first introduce a query function $g(Q, d)$. The definition of function is shown in Algorithm~\ref{alg:query}.
For a given rank $d$, the function returns a $x$ whose rank is close to $d$. This property is formally described in the following Lemma.
\begin{thm:lemma}\label{lem:query}
For a given $\eps$-approximate summary $Q(\sD) = (S, \tdr^+, \tdr^-, \tdw)$, $x^*=g(Q,d)$ satisfies the following property
\begin{equation}
\begin{split}
    d &\geq \tdr^+(x^*) - \tdw(x^*) - \frac{\eps}{2} \w(\sD)\\
    d &\leq \tdr^-(x^*) +\tdw(x^*) + \frac{\eps}{2} \w(\sD)
\end{split}
\end{equation}
\end{thm:lemma}
\begin{proof}
We need to discuss four possible cases
\begin{itemize}
\item $d < \frac{1}{2}[\tdr^-(x_1) + \tdr^+(x_1) ]$ and $x^* = x_1$. Note that the rank information for $x_1$ is accurate
($\tdw(x_1) = \tdr^+(x_1) = \w(x_1)$, $\tdr^-(x_1)=0$), we have
\begin{equation*}
\begin{split}
 d &\geq 0 - \frac{\eps}{2} \w(\sD) =  \tdr^+(x_1) - \tdw(x_1) - \frac{\eps}{2} \w(\sD) \\
 d &< \frac{1}{2}[\tdr^-(x_1) + \tdr^+(x_1) ]\\
 &\leq \tdr^-(x_1) + \tdr^+(x_1)\\
 &= \tdr^-(x_1) + \tdw^+(x_1)
\end{split}
\end{equation*}
\item $d \geq \frac{1}{2}[\tdr^-(x_k) + \tdr^+(x_k) ]$ and $x^* = x_k$, then
\begin{equation*}
\begin{split}
 d &\geq \frac{1}{2}[\tdr^-(x_k) + \tdr^+(x_k) ]\\
 & = \tdr^+(x_k) - \frac{1}{2}[\tdr^+(x_k) - \tdr^-(x_k)] \\
   &= \tdr^+(x_k) - \frac{1}{2} \tdw(x_k)\\
 d &< \w(\sD) + \frac{\eps}{2} \w(\sD) = \tdr^-(x_k) +\tdw(x_k) + \frac{\eps}{2} \w(\sD)
\end{split}
\end{equation*}
\item $x^* = x_i$ in the general case, then
\begin{equation*}
\begin{split}
2 d  &<  \tdr^-(x_i) + \tdw(x_i) + \tdr^+(x_{i+1}) - \tdw(x_{i+1})\\
&= 2[\tdr^-(x_i) + \tdw(x_i)] + [\tdr^+(x_{i+1}) - \tdw(x_{i+1}) - \tdr^-(x_i) - \tdw(x_i)]\\
&\leq 2[\tdr^-(x_i) + \tdw(x_i)] + \eps \w(\sD)\\
 2 d &\geq \tdr^-(x_i) + \tdr^+(x_i)\\
 & = 2[ \tdr^+(x_i) - \tdw(x_i)] - [\tdr^+(x_i) - \tdw(x_i) - \tdr^-(x_i) ] + \tdw(x_i)\\
 &\geq 2[ \tdr^+(x_i) - \tdw(x_i)] - \eps \w(\sD) + 0
\end{split}
\end{equation*}
\item $x^*=x_{i+1}$ in the general case
\begin{equation*}
\begin{split}
2 d  &\geq  \tdr^-(x_i) + \tdw(x_i) + \tdr^+(x_{i+1}) - \tdw(x_{i+1})\\
&= 2[\tdr^+(x_{i+1}) - \tdw(x_{i+1})] \\
& \ \ \ \ -  [\tdr^+(x_{i+1}) - \tdw(x_{i+1}) - \tdr^-(x_i) - \tdw(x_i)]\\
&\geq 2[\tdr^+(x_{i+1}) + \tdw(x_{i+1})] - \eps \w(\sD)\\
 2 d & \leq \tdr^-(x_{i+1}) + \tdr^+(x_{i+1})\\
 & = 2[ \tdr^-(x_{i+1}) + \tdw(x_{i+1})] \\
 & \ \ \ \ + [\tdr^+(x_{i+1}) - \tdw(x_{i+1}) - \tdr^-(x_{i+1}) ] - \tdw(x_{i+1})\\
 &\leq 2[ \tdr^-(x_{i+1}) + \tdw(x_{i+1})]  + \eps \w(\sD) - 0
\end{split}
\end{equation*}
\end{itemize}
\end{proof}
Now we are ready to introduce the prune operation. Given a quantile summary $Q(\sD) = (S, \tdr^+, \tdr^-, \tdw)$ with $S=\{x_1,x_2,\cdots, x_k\}$ elements, and a memory budget $b$. The prune operation creates another summary $Q'(\sD) = (S', \tdr^+, \tdr^-, \tdw)$ with $S' =\{x'_1,x'_2, \cdots, x'_{b+1}\}$, where $x'_{i}$ are selected by query the original summary such that
$$ x'_{i} = g\left(Q, \frac{i-1}{b}\w(\sD)\right).$$
The definition of $\tdr^+, \tdr^-, \tdw$ in $Q'$ is copied from original summary $Q$, by restricting input domain from $S$ to $S'$.
There could be duplicated entries in the $S'$. These duplicated entries can be safely removed to further reduce the memory cost.
Since all the elements in $Q'$ comes from $Q$, we can verify that $Q'$ satisfies all the constraints in Definition~\ref{def:sketch} and is a valid quantile summary.

\begin{thm:thm}
Let $Q'(\sD)$  be the summary pruned from  an $\eps$-approximate quantile summary $Q(\sD)$ with $b$ memory budget.
 Then $Q'(\sD)$ is a $(\eps+\frac{1}{b})$-approximate summary.
\end{thm:thm}
\begin{proof}
We only need to prove the property in Eq.~\eqref{eq:constraint2} for $Q'$.
Using Lemma~\ref{lem:query}, we have
\begin{equation*}
\begin{split}
    \frac{i-1}{b}\w(\sD)   + \frac{\eps}{2} \w(\sD)  &\geq \tdr^+(x'_i) - \tdw(x'_i)\\
    \frac{i-1}{b}\w(\sD)   - \frac{\eps}{2} \w(\sD) &\leq \tdr^-(x'_i) +\tdw(x'_i)
\end{split}
\end{equation*}
Combining these inequalities gives
\begin{equation*}
\begin{split}
 &\tdr^+(x'_{i+1}) - \tdw(x'_{i+1})  -  \tdr^-(x'_{i}) - \tdw(x'_{i}) \\
 \leq& [ \frac{i}{b}\w(\sD) + \frac{\eps}{2} \w(\sD)] - [ \frac{i-1}{b}\w(\sD) - \frac{\eps}{2} \w(\sD)] = (\frac{1}{b} + \eps)\w(\sD)
\end{split}
\end{equation*}
\end{proof}

% That's all folks!
\end{document}